\pdfoutput=1

\documentclass[11pt]{article}

\usepackage[final]{acl}

\usepackage{times}
\usepackage{latexsym}

\usepackage[T1]{fontenc}

\usepackage[utf8]{inputenc}

\usepackage{microtype}

\usepackage{inconsolata}

\usepackage{graphicx}

\usepackage{inconsolata}

\usepackage{hyperref}
\usepackage{transparent}
\usepackage{array}
\usepackage{multirow}
\usepackage{multicol}
\usepackage{xprintlen}
\usepackage{csquotes}
\usepackage{enumitem}
\usepackage[normalem]{ulem}
\usepackage[export]{adjustbox}
\usepackage{float}
\usepackage{placeins}
\usepackage{amsmath}
\usepackage{makecell}
\usepackage{colortbl}
\usepackage{siunitx}
\usepackage{ragged2e}

\newcolumntype{P}[1]{>{\centering\arraybackslash}m{#1}}

\definecolor{darkgreen}{HTML}{14781c}

\newcommand{\cancut}[1]{}   %

\definecolor{babyblue}{rgb}{0.54, 0.81, 0.94}
\definecolor{peach}{rgb}{.9, .6, .36}
\definecolor{lightgreen}{rgb}{.37, .63, .47}

\title{MASIVE: Open-Ended Affective State Identification in English and Spanish}

\author{
 \textbf{Nicholas Deas}, %
 \textbf{Elsbeth Turcan}, %
 \textbf{Iv\'{a}n P\'{e}rez Mej\'{i}a}, %
 \textbf{Kathleen McKeown} %
\\
\\
Columbia University, Department of Computer Science
\\
 \small{
   \textbf{Correspondence:} \href{mailto:ndeas@cs.columbia.edu}{[ndeas,eturcan,iep2110,kathy]@cs.columbia.edu}
 }
}

\begin{document}
\maketitle

\begin{abstract}

In the field of emotion analysis, much NLP research focuses on identifying a limited number of discrete emotion categories, often applied across languages. 
These basic sets, however, are rarely designed with textual data in mind,
and
culture, language, and dialect can influence how particular emotions are interpreted.
In this work, we broaden our scope 
to a practically unbounded set of \textit{affective states}, which includes any terms that humans use to describe their experiences of feeling. We collect and publish 
MASIVE, a dataset of Reddit posts in English and Spanish containing over 1,000 unique affective states each.
We then define the new problem of \textit{affective state identification} for language generation models framed as a masked span prediction task. On this task, we find that smaller fine-tuned multilingual models 
outperform much larger LLMs, even on region-specific Spanish affective states. Additionally, we show that pre-training on MASIVE improves model performance on existing emotion benchmarks. Finally, through machine translation experiments, we find that native speaker-written data is vital to good performance on this task.

\end{abstract}

\section{Introduction}

In the field of emotion analysis, much NLP research focuses on identifying a limited number of discrete emotion categories, typically using \textit{basic emotion sets} from the field of psychology \citep{plaza-del-arco-etal-2024-emotion-analysis}. These basic emotion sets are rarely designed with textual expression in mind (e.g., \citealt{ekman-basic}, whose model defines basic emotions by the recognizability 
of 
facial expressions), and very little research examines the validity of 
adapting these sets to textual data.

Emotion analysis furthermore relies on largely the same emotion categories across languages, including, in some cases, translating resources such as lexicons,
fine-tuning data, or evaluation data \cite{dhananjaya-lexicon,isbister-translation,kathunia-translation}. Previous research has also shown that existing multilingual models encode meaning in an Anglocentric way \citep{havaldar-multilingual}. As 
recent studies have found that culture and language 
influence the meaning of emotional terms like "love" \cite{jackson-semantics}, models that fail to understand cultural context or rely on 
mainstream dialects may also fail to capture the nuances of an author's 
expression \citep{deas-aal}.

\begin{figure}
    \centering
    \includegraphics[width=.4\textwidth, trim=0.8cm 1cm 0.8cm 0.8cm]{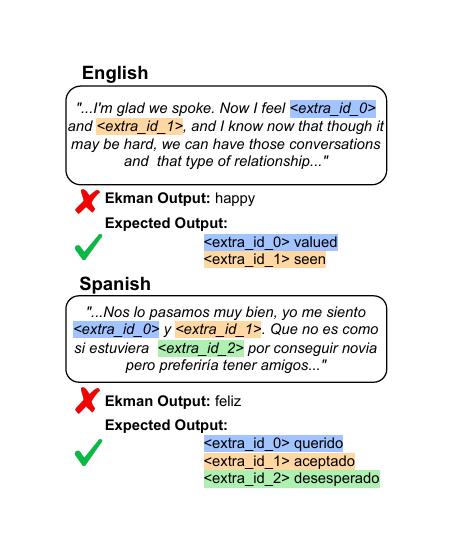}
    \caption{Paraphrased input and expected output examples from MASIVE in English and Spanish. Models are tasked with predicting \textit{affective states} (highlighted), which reflect more nuanced feelings than label sets in prior work, such as the Ekman basic emotions.}
    \label{fig:data-example}
\end{figure}

In this work, we argue for a descriptive approach to emotion analysis. We broaden our scope from a small set of 
basic emotions to a practically unbounded set of \textit{affective states} \citep{vandenbos-apa}, which includes any terms that humans use to describe their experiences of feeling, including emotions, moods, and figurative expressions of feelings (e.g. "blue" as an
expression of sadness instead of the color). 
We then define the new problem of \textit{affective state identification} (ASI), which is a targeted masked span prediction task: given a text description of an emotional experience, we train models to produce single-word affective states that correspond to the description. These affective states may include common emotion categories such as \textit{happy} or \textit{sad}, but they also allow us to incorporate nuance and intensity (e.g., \textit{elated}, \textit{calm}, \textit{jealous}, \textit{lonely}, etc.) as well as other classifications that are not typically considered emotions such as moods (e.g., longer-term feelings of being \textit{motivated} or \textit{stuck}). 

We collect MASIVE: Multilingual Affective State Identification with Varied Expressions, a new benchmark dataset for ASI using Reddit data. We use a bootstrapping procedure to discover new affective state labels and collect posts containing natural emotional expressions in English and Spanish, yielding 1600 unique affective state labels in English and 1000 in Spanish.
We evaluate 
our data collection methods
with human annotation, finding that 88\% and 72\% of our automatically collected English and Spanish labels, respectively,
reflect affective states, and document unique features of the data including negations and, in Spanish, grammatical gender.
We then use this dataset to evaluate the performance of 
commonly-used generative models,
finding that small fine-tuned models generally outperform LLMs. Beyond ASI, we experiment with using our corpora as pre-training data and show that 
MASIVE incorporates
knowledge that generalizes to existing emotion detection benchmarks. Finally, we assess fine-tuning and evaluating models on machine-translated data
and find that original texts written by native speakers are essential for performing 
ASI.

Our contributions in this work are as follows: 

\begin{enumerate}[noitemsep, topsep=0pt]
    \item We introduce a novel benchmark for ASI with language generation models, including a significantly larger label set than prior related benchmarks\footnote{We make our code and data available at \url{https://github.com/NickDeas/MASIVE}};
    \item We benchmark multilingual models and show that smaller, fine-tuned models outperform current LLMs on this dataset; 
    \item We analyze the behavior and performance of models on region-specific affective language, grammatical gender, and negations; and
    \item We empirically argue that both fine-tuning and evaluating on texts authored by native speakers is vital for capturing nuances in multilingual affective writing
\end{enumerate}

\section{Affective State Identification (ASI)}
In contrast to traditional emotion detection, we propose a novel task, ASI. Our goal is to capture the broad set of ways in which humans describe their own feelings in text. We refer to these expressions as \textit{affective states} \cite{vandenbos-apa}; 
this is an umbrella term incorporating multiple kinds of feelings such as emotions and moods. 

We highlight multiple implications of models that are capable of accurately inferring affective states. First, ASI enables the identification of more nuanced feelings derived from textual expressions of affect (e.g., distinguishing \textit{despair} and \textit{grief} which may be similarly described by basic emotions). Additionally, ASI-capable models can be adapted to multiple theories of emotion. Whereas a model specifically trained on one set of emotions (e.g., Ekman) must be fine-tuned for each label set (e.g., 27 emotions of \citealt{cowen-distinct}), ASI models can be restricted to an arbitrary subset of affective state labels. Finally, ASI is grounded in expressions of feelings by the author in contrast to perceived emotion labels generated by annotators. These perceived emotion labels may train models to encode spurious factors affecting human emotion perception, such as cultural differences.

\section{Data}

\subsection{MASIVE Corpus}
\begin{figure}[!htbp]
    \centering
    \includegraphics[width=1.0\linewidth,trim={1cm 0 0cm 0},clip]{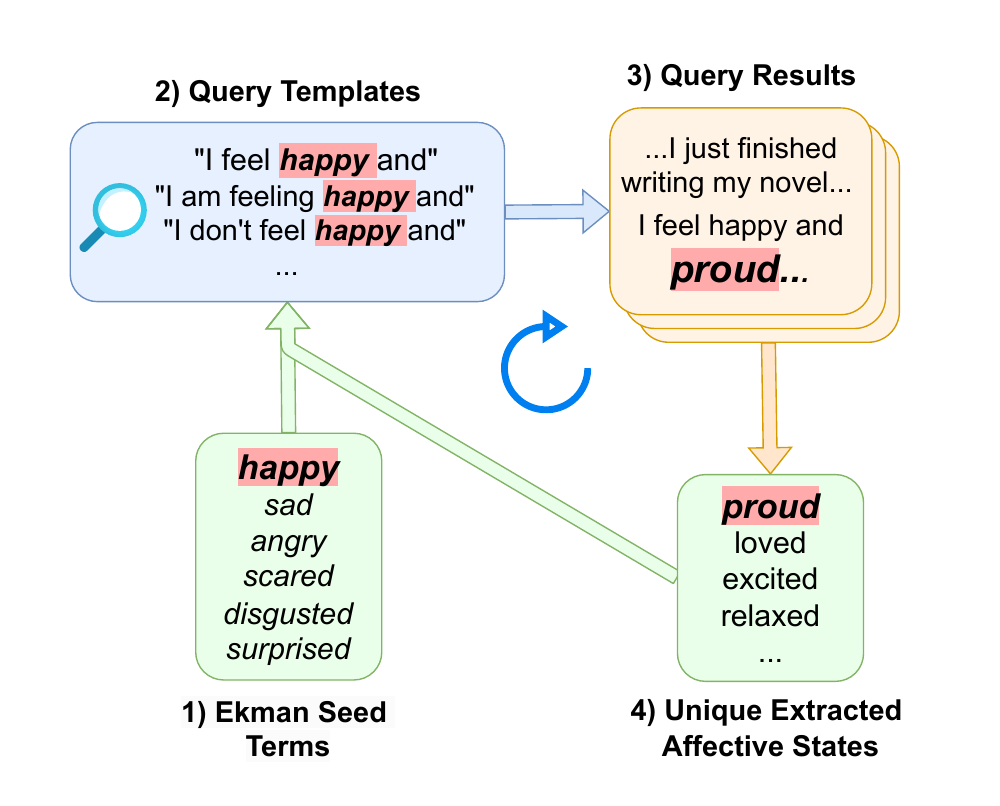}
    \caption{Illustration of the bootstrapping procedure used to collect texts and automatically extracted affective state labels in the MASIVE corpus.}
    \label{fig:bootstrap-diag}
\end{figure}
We collect texts with expressions of affective states from Reddit\footnote{Using the PullPush API at \url{https://pullpush.io/}} using a bootstrapping procedure (illustrated in \autoref{fig:bootstrap-diag}). 
We source data from Reddit due to the availability of large quantities of diverse texts, and because many submissions reflect casual narratives where authors are likely to express their feelings. Beginning with the adjective forms of the Ekman emotions (Step 1 in \autoref{fig:bootstrap-diag}), we search for texts containing forms of ``\textit{I feel <affect> and}...'', ``\textit{I am feeling <affect> and}...'', where \textit{<affect>} is replaced with each emotion term (Step 2). Notably, we also search for ``\textit{I don't feel <affect> and}...'' and ``\textit{I am not feeling <affect> and}...'' to better capture the diversity of ways in which authors can
express feelings.
We extract affective state terms that follow the ``\textit{and}'' 
from the retrieved posts (Step 3)
to form a new set of search phrases with these terms (Step 4). We repeat these steps, expanding the pool of query affective states in each round. Our primary assumption is that any adjective conjuncts of the 
query emotion term are also affective states, regardless of whether they are canonical emotion terms. For example, if "happy" was used to query the text "I feel happy and excited," the term "excited" is both an adjective and a conjunct; the same is true of ``light'' in ``I feel happy and light''. In contrast, in 
"I feel happy and want to smile", "want" is a verb and would not be considered an affective state. We evaluate 
this assumption
in \autoref{sec:data-val}.
 
In Spanish, we conduct the same procedure using forms of ``\textit{Estoy <affect> y}...'', ``\textit{Me siento <affect> y}...'', and ``\textit{Estoy sintiendo <affect> y}...''.
We also seed the process with the most common Spanish translations of the Ekman emotions 
on Reddit (see \autoref{app:data}). Additionally, as Spanish includes both masculine and feminine forms for some terms, we search for both forms where applicable.
Finally, we also collect a challenge set including affective state labels associated with regional Spanish varieties, hand-selected by a native Spanish-speaker, to evaluate models' abilities to generalize to 
less-represented
dialects (see \autoref{app:data}). 

For both English and Spanish, we run 4 rounds of bootstrapping;
for the regional Spanish terms, we run only a single round to avoid introducing non-regional terms.
15 affective states were randomly sampled from both datasets, and all posts containing those 15 affective states were reserved as part of each test set to evaluate models on unseen affective states. Summary statistics describing the English and Spanish splits as well as the regional Spanish challenge set are included in \autoref{tab:data-stats}. We include examples of samples in MASIVE in \autoref{app:data-ex}, and following \citet{bender-data}, we include a Data Statement in \autoref{app:data-statement}.

\begin{table}[ht]
    \centering
    \small
    \addtolength{\tabcolsep}{-0.2em}
    \begin{tabular}{ P{.6cm}  P{.7cm} | P{1.0cm} P{1.0cm} P{1.1cm} P{1.2cm} }
        \hline
        Lang                & Split     & Size      & Input Length  & \# AS/ Text & \# Unique AS\\
        \hline
        \multirow{3}{*}{En} & Train & 93,736 & 310.99 & 1.11 & 1,627 \\
                            & Test & 10,049 & 306.61 & 1.13 & 775 \\
                            & Chal & 4,720 & 299.00 & 1.15 & 320 \\
        \hline
        \multirow{4}{*}{Es} & Train & 30,958 & 240.99 & 1.06 & 1,002 \\
                            & Test & 4,274 & 247.90 & 1.08 & 618 \\
                            & Chal & 1,557 & 245.50 & 1.12 & 145 \\
                            & Reg & 559 & 233.95 & 1.07 & 59 \\
        \hline
    \end{tabular}
    \caption{Summary statistics of English and Spanish MASIVE. Text lengths are measured in mT5 tokens. AS = Affective State; Chal = unseen challenge set}
    \label{tab:data-stats}
\end{table}

\subsection{Data Validation Procedure}
\label{sec:data-val}

To validate the assumptions of our bootstrapping procedure and examine 
how affective states are used in our dataset, we collect human evaluations of 
the automatically identified affective states. 
Judgments are conducted by 2 native Spanish-speakers in Iberian Culture studies and 
2 native English-speakers in Psychology for Spanish and English respectively. We randomly sample 200 texts from each language's test set
for evaluation such that 50 texts are shared by each pair. Annotators are provided a full Reddit post with a single automatically-identified affective state highlighted.

    \begin{table*}[!htbp]
        \centering
        \small
        \begin{tabular}{ P{5cm} | l }
            \hline
            \textbf{Category} & \textbf{Example Affective States} \\
            \hline
            Basic Emotions and Alternatives & \makecell*[l]{\textbf{Ekman:}  happy, sad, surprised, angry, afraid, disgusted \\ \textbf{Plutchik:} remorseful, bored, ecstatic, etc. \\ \textbf{Alternatives: } (\textit{trust}) safe, comfortable, protected\\ \qquad \qquad \qquad (\textit{anticipation}) ready, excited, hopeful} \\
            \hline
            Variation in Intensity & \makecell*[l]{\textbf{happy}: giddy, euphoric, ecstatic, content, joyful \\ \textbf{angry}: grumpy, enraged, resentful, irritated \\ \textbf{afraid}: terrified, frightened, alarmed, fearful \\ \textbf{surprised}: shocked,  amazed, dazed, stunned }\\
            \hline
            Context-Dependent & disconnected, uplifted, gullible, disrespected, underwhelmed \\
            \hline
            Metaphorical/Figurative Usage & weak, nauseous, empty, lost, spent \\
            \hline
        \end{tabular}
        \caption{Examples of different types of discovered English affective states manually categorized by the authors. }
        \label{tab:as-ex}
    \end{table*}

    \begin{table}[ht]
        \centering
        \begin{tabular}{ c | P{.8cm} P{.8cm} P{.9cm} | P{.8cm} P{.8cm} }
            \hline
            \multirow{2}{*}{\centering Lang} & \multicolumn{3}{c|}{Human-Annotated} & \multicolumn{2}{c}{Automatic} \\
            & Aff & Fig & Emo &  Neg & Fem \\
            \hline
            En   & 88.4\% & 58.8\% & 34.2\% & 7.75\% & \cellcolor{gray} \\
            Es   & 71.5\% & 38.5\% & 18.5\% & 27.0\% & 28.0\% \\
            \hline
        \end{tabular}
        \caption{Human and automatic analysis of how affective states in the English and Spanish datasets are used in context. \textit{Aff}: Affective State; \textit{Fig}: Figurative; \textit{Emo}: Emotion; \textit{Neg}: Negations; \textit{Fem}: Feminine Form}
        \label{tab:data-val}
    \end{table}

We ask annotators to judge the term in context on 3 dimensions,
beginning with whether the highlighted term reflects an affective state.
If a term is judged to reflect an affective state, annotators are asked to judge whether the highlighted term better reflects an emotion or a mood\footnote{We distinguish emotions -- shorter-term feelings triggered by identifiable events -- from moods -- longer-term feelings not necessarily triggered by an event.} and whether the highlighted term is used figuratively (e.g., "blue") or literally (e.g., "sad").
All 3 dimensions are judged on 4-point Likert scales where higher values mean the term primarily reflects an affective state, an emotion, and a literal usage, respectively. According to Cohen's kappa, annotators achieved moderate agreement in English ($\kappa=.51$) and substantial agreement in Spanish ($\kappa=.69$). Additional details concerning human annotations are included in \autoref{app:data}.

Additionally, we analyze 2 aspects of our dataset that differentiate it from prior emotion detection benchmarks. First, because Spanish is a language with grammatical gender for adjectives, part of the affective state prediction problem in MASIVE includes choosing whether to use the masculine or feminine form in the context of the input. Second, authors in natural settings may also tend to express their feelings by stating how they do \textit{not} feel (e.g., ``I'm not happy, but....''), and we specifically include negations to test models' capability to contend with this construction in both English and Spanish.

\subsection{Data Analysis}

The results of the aforementioned data annotations as well as automatic statistics are included in \autoref{tab:data-val}. Human annotation results are reported as the percentage of affective states within the sample;
for negations and grammatical gender, we report the percentage of texts in our datasets that include any target negations or any feminine adjectives.\footnote{Recall that a single datapoint may have multiple labels joined by \textit{and}.} Large majorities (88\% and 72\% in English and Spanish respectively) of terms were judged to reflect affective states, validating the contents of MASIVE. Additionally, annotators identified most affective states as longer-term moods without specific triggers (65.8\% and 81.5\% respectively). Finally, a significant portion of texts in both languages were determined to reflect figurative use rather than terms that are typically affective states (58.8\% and 38.5\% respectively), presenting a unique challenge to models compared to prior work.

    We broadly categorize different types of affective states included in MASIVE that distinguish ASI from prior emotion detection work in \autoref{tab:as-ex}. Beyond the seed Ekman emotions, MASIVE also contains many emotions from other models, such as the Plutchik emotion wheel or those listed in \citet{wang-emotion}. Emotions not included tend to lack clear adjective forms (e.g., \textit{trust}), but MASIVE does include alternatives of similar meaning (e.g., \textit{safe}, \textit{comfortable}). Many emotions capture variations in intensity of basic emotions, such as \textit{euphoric} or \textit{content} which, in part, reflect different intensities of \textit{happy}. In contrast to prior emotion detection work, our bootstrapping procedure captures many affective states that depend on a particular social context, such as \textit{underwhelmed} which specifically implies unmet expectations.

\subsection{Fixed-Label Set Data} 
\label{sec:external-data}

We additionally evaluate the performance of MASIVE-fine-tuned models on two previously published datasets in both English and Spanish. A key distinction from MASIVE is that these datasets feature limited label sets; we describe our evaluation procedures in \autoref{sec:metrics}. In English, 
we evaluate on GoEmotions \citep{demszky-goemo}, a commonly-used emotion dataset consisting of Reddit comments; it is originally labeled with 27 distinct emotion categories, though the authors also relabel the data 
with the Ekman basic emotions. We additionally evaluate on EmoEvent \citep{plaza-del-arco-emoevent}, a dataset with both English and Spanish subsets of Tweets (among other languages) also labeled with the Ekman set.

\subsection{Machine-Translated Data} 
Finally, we conduct two cross-lingual experiments expanding on prior work investigating the use of machine translation 
for cross-lingual generalization \citep{isbister-translation,kathunia-translation}. In contrast to prior findings, however, we hypothesize that neither translating the training nor evaluation data will result in competitive performance with models trained on native data. 

We evaluate both \textit{translate-train} and \textit{translate-test} approaches \cite{hu-xtreme}. First, translate-train involves translating the training data from a source language (e.g., English) into a target language (e.g., Spanish). A model is then subsequently fine-tuned on the generated training data in the target language (e.g., English translated into Spanish). 
Alternatively, translate-test leverages a model already trained in a source language (e.g., English). 
Test data texts in the target language (e.g., Spanish) are translated into the source language (e.g., Spanish translated into English) at inference time.

In both settings, we use bilingual Opus-MT 
models \citep{tiedmann-opusmt} 
to independently translate the input documents and target affective state labels. We select Opus-MT models following \citet{kathunia-translation} and because they are accessible, open-source models, reflecting resources that may be used for large scale translation.
Models fine-tuned on translated data or subset data are denoted $_{Tr}$ and $_{S}$ respectively, and translated test sets are also denoted $_{Tr}$.

\section{Experimental Configuration}

\subsection{Models}

We experiment with fine-tuning small language models on our original and machine-translated data.
We also perform experiments with two Large Language Models (LLMs) 
in a zero-shot setting.

\paragraph{fine-tuned Generative Models.} 
Most of our models are based on mT5-Large \citep{xue-etal-2021-mt5}.
During fine-tuning and prediction on 
MASIVE, we mask affective state 
words 
wherever they appear and task models to fill them,
mimicking mT5's pre-training. We additionally experiment with T5-large \citep{raffel-t5} for English only\footnote{No comparable monolingual T5 checkpoint for Spanish has been made publicly available.}. In the results, models' superscripts denote that a model was fine-tuned on our English (T5$^{En}$ and mT5$^{En}$) or Spanish 
(mT5$^{Es}$) corpus.

\paragraph{Large Language Models. } We evaluate two modern, open-source LLMs--Llama-3\footnote{\url{https://llama.meta.com/llama3}}
and Mixtral-Instruct \citep{jiang-mixtral}--as these models have been specifically evaluated 
in multilingual settings. 
We instruct these models 
to perform the same masked token prediction task as mT5 (see \autoref{app:methods}). Due to context window constraints and input lengths, LLMs are evaluated in a zero-shot setting. Further checkpoint and generation hyperparameter details are included in \autoref{app:methods}.

\subsection{Metrics} \label{sec:metrics}

\subsubsection{MASIVE Evaluation} We report \textbf{top-k accuracy}  
for our models with $k \in \{1, 3, 5\}$\footnote{As some samples in the datasets have multiple labels, we calculate top-k accuracy at the sample level using beam search and report average sample-level scores.}, along with two generative metrics: the \textbf{negative log-likelihood} (NLL) of the gold affective state and the model's \textbf{log perplexity}. 
In Spanish, if the gendered form of the prediction does not match that of the gold term (e.g. enojado vs. enojada), the prediction is considered incorrect, but the similarity of the prediction in these cases is captured by the \textbf{top-k similarity} metric, which we describe below.

\paragraph{Top-k Similarity.} Because our label set is very large, we also report a measure of similarity between the model's top predictions and the gold. Here, we rely on contextual embeddings using 
multilingual, pre-trained BERT-base \citep{devlin-bert}.
To ensure that the similarity model encodes 
affective senses of each term, we embed
the 
predicted and gold
emotion terms within 100-token contexts from the original post and calculate cosine similarity between them. 
We report the maximum similarity of these contextual embeddings when looking at the top 1, 3, and 5 most likely model predictions. Full details are available in \autoref{app:topk-sim}.

\begin{table*}[ht]
    \centering
    \resizebox{\textwidth}{!}{%
    \begin{tabular}{ll|rr|rrr|rrr}
        \hline
        Lang & Model                       & \multicolumn{1}{l}{NLL$\downarrow$} & \multicolumn{1}{l|}{Log Perp$\downarrow$} & \multicolumn{1}{l}{Acc@1$\uparrow$} & \multicolumn{1}{l}{Acc@3$\uparrow$} & \multicolumn{1}{l|}{Acc@5$\uparrow$} & \multicolumn{1}{l}{Sim@1$\uparrow$} & \multicolumn{1}{l}{Sim@3$\uparrow$} & \multicolumn{1}{l}{Sim@5$\uparrow$} \\ \hline
        \multirow{4}{*}{En} & T5$^{En}$	 & 6.87   & 6.85              & 20.05\% & 29.44\% & 34.64\% & 0.569 & 0.673 & 0.718 \\
                            & mT5$^{En}$ & 10.93  & 10.90             & \textbf{17.91\%} & \textbf{26.81\%} & \textbf{30.90\%} & \textbf{0.564} & \textbf{0.670} & \textbf{0.711} \\
                            & Llama-3	 & 60.79  & 44.84             & 1.29\%  & 2.26\%  & 2.92\%  & 0.431 & 0.460 & 0.475 \\
                            & Mixtral	 & 1.52   & \cellcolor{gray}  & 7.83\%  & 8.93\%  & 10.55\% & 0.475 & 0.495 & 0.518 \\
        \hline
        \multirow{3}{*}{Es} & mT5$^{Es}$ & 6.91   & 6.89              & \textbf{24.51\%} & \textbf{36.16\%} & \textbf{41.23\%} & \textbf{0.610} & \textbf{0.734} & \textbf{0.781} \\
                            & Llama-3	 & 77.78  & 61.56             & 2.52\%  & 4.69\%  & 5.91\%  & 0.445 & 0.480 & 0.498 \\
                            & Mixtral	 & 1.47   & \cellcolor{gray}  & 16.80\% & 19.47\% & 22.24\% & 0.525 & 0.553 & 0.583 \\
        \hline
    \end{tabular}
    }%
    \caption{Comparison of T5, mT5, and two LLMs on our proposed Reddit dataset, aggregated scores only. Note that the Spanish test set and the English test set are not directly comparable as noted in \autoref{sec:metrics}. \textbf{Bolded} scores highlight the best-performing multilingual model.
    }
    \label{tab:internal-eval}
\end{table*}

\begin{table*}[ht]
    \adjustbox{max width=\textwidth}{%
    \centering
    \small
    \begin{tabular}{l|l|rrr|lll|lll}
        \hline
        Dataset                                          & Model & \multicolumn{1}{c}{P} & \multicolumn{1}{c}{R} & \multicolumn{1}{c}{F1} & Acc@1& Acc@3 & Acc@5 & Sim@1 & Sim@3 & Sim@5 \\ 
        \hline
        \multicolumn{1}{l|}{\multirow{2}{*}{GoEmotions (7)}}    
                                                        & mT5	         & \textbf{33.63} & 19.28 & 16.25 & \textbf{38.49\%} & \textbf{70.73\%} & \textbf{85.99\%} & \textbf{0.736} & \textbf{0.884} & \textbf{0.946} \\
                                                        & mT5$^{MAS}$	 & 33.06 & \textbf{39.81} & \textbf{28.30} & 17.49\% & 32.11\% & 39.25\% & 0.629 & 0.733 & 0.771 \\
        \hline
        \multicolumn{1}{l|}{\multirow{2}{*}{GoEmotions (27)}}   
                                                        & mT5	         & 12.57 & 4.77  & 2.24  & 2.53\% & \textbf{12.90\%} & \textbf{23.51\%} & \textbf{0.525} & \textbf{0.614} & \textbf{0.670} \\
                                                        & mT5$^{MAS}$	 & \textbf{27.08} & \textbf{18.76} & \textbf{11.92} & \textbf{7.54\%} & 12.22\% & 15.16\% & 0.508 & 0.602 & 0.639 \\
        \hline
        \multicolumn{1}{l|}{\multirow{2}{*}{EmoEvent (En)}}	    
                                                        & mT5	         & 30.06 & 14.36 & 2.84  & 10.50\% & \textbf{71.70\%} & \textbf{93.64\%} & 0.630 & \textbf{0.880} & \textbf{0.974} \\
                                                        & mT5$^{MAS}$	 & \textbf{34.81} & \textbf{32.74} & \textbf{29.55} & \textbf{33.40\%} & 57.06\% & 69.38\% & \textbf{0.712} & 0.842 & 0.893 \\
        \Xhline{4\arrayrulewidth}
        \multicolumn{1}{l|}{\multirow{2}{*}{EmoEvent (Es)}}	    
                                                        & mT5	         & 26.13 & 14.52 & 6.41  & 24.29\% & 70.34\% & \textbf{89.12\%} & 0.713 & 0.882 & \textbf{0.955} \\
                                                        & mT5$^{MAS}$	 & \textbf{54.93} & \textbf{21.54} & \textbf{17.80} & \textbf{39.75\%} & \textbf{82.62\%} & 86.11\% & \textbf{0.750} & \textbf{0.918} & 0.935 \\
        \hline
    \end{tabular}
    }%
    \caption{Performance of mT5 fine-tuned on emotion classification datasets, with and without prior fine-tuning on MASIVE. \textbf{Bolded} scores highlight the best performing model on each dataset under each metric.}
    \label{tab:fixed-label-english}
\end{table*}

\subsubsection{Fixed-Label Set Evaluation} \label{sec:external-eval-methods}
To evaluate how well our dataset imbues models with general emotional knowledge, 
we evaluate two variants of mT5: first, mT5 fine-tuned only on existing emotion benchmarks, and second, mT5 fine-tuned on MASIVE followed by existing benchmarks (denoted with superscript $^{MAS}$).

We frame affective state detection as a generative mask-filling task
rather than
a classification task. Therefore, 
to adapt the evaluation sets to our generative setting, we 
append "I feel <extra\_id\_0>" to the end of each input 
to match the format of our evaluation on MASIVE (see \autoref{fig:data-example}), using adjective forms of the 
gold emotion labels. In this setting, we report \textbf{top-k accuracy} and \textbf{similarity} as we do for 
MASIVE.
Additionally, to adapt models to a fixed-label set, we sort the fixed set of emotion labels by their likelihood according to the model and select the most probable emotion label
as the prediction. For these experiments, we report macro \textbf{precision}, \textbf{recall}, and \textbf{F1 score}.

\section{Results}

\subsection{MASIVE Evaluation}

\autoref{tab:internal-eval} presents the performance metrics for fine-tuned mT5, Llama-3, and Mixtral on 
our English and Spanish test sets, as well as fine-tuned T5
for the English test set only. Among multilingual
models, \textbf{fine-tuned mT5 outperforms 
both LLMs on top-k accuracy and top-k similarity for both languages (Takeaway \#1)}, despite having drastically fewer parameters.\footnote{Llama-3 occasionally refuses to make a prediction if the content discussed is sensitive (e.g., drug use). Results taking invalid responses into account are included in \autoref{app:results}.}
Between the LLMs, Mixtral outperforms Llama-3. This performance difference may be explained by the difference in size between models, as well as the fact that multilingual data was upsampled in Mixtral's pre-training compared to prior models. 

In English,
the large variant of T5 has been shown to slightly outperform mT5 \cite{xue-mt5}. 
We find a similar difference,
and in fact, monolingual T5 
outperforms all other models in English.
Because the remaining experiments include Spanish data, we focus on mT5. We note, however, that \textbf{dedicated monolingual
models may offer significantly higher performance on ASI (Takeaway \#2)} and leave further exploration of the differences between monolingual and multilingual models to future work. 

While the differences in language and content of the English and Spanish datasets prevent us from making conclusions concerning their relative difficulty, \autoref{tab:internal-eval} also shows that performance in Spanish tends to be higher than in English, 
despite the better representation of English in pre-training and larger size of the collected English data compared to Spanish. This trend could be due to the larger set of unique affective states in our English data than Spanish, with more nuanced affective states that may be difficult for models to predict accurately.

\begin{table*}[ht]
    \centering
    \small
    \resizebox{\textwidth}{!}{%
        \begin{tabular}{l|l|l|rrr|rrr}
        \hline
        Lang & Model & Subset                       & \multicolumn{1}{l}{Acc@1$\uparrow$} & \multicolumn{1}{l}{Acc@3$\uparrow$} & \multicolumn{1}{l|}{Acc@5$\uparrow$} & \multicolumn{1}{l}{Sim@1$\uparrow$} & \multicolumn{1}{l}{Sim@3$\uparrow$} & \multicolumn{1}{l}{Sim@5$\uparrow$} \\ 
        \hline
        \multirow{4}{*}{\centering En} & \multirow{2}{*}{T5$^{En}$}	 & Seen   & 35.22\% & 50.31\% & 57.70\% & 0.640 & 0.756 & 0.805 \\
                                       &        	                  & Unseen & 2.92\%  & 5.88\%  & 8.61\%  & 0.488 & 0.579 & 0.620 \\
        \cline{2-9}
                                       & \multirow{2}{*}{mT5$^{En}$} & Seen   & 32.85\% & 48.90\% & 56.13\% & 0.633 & 0.757 & 0.804 \\
                                       &        	                  & Unseen & 1.04\%  & 1.88\%  & 2.41\%  & 0.487 & 0.571 & 0.607 \\
        \Xhline{4\arrayrulewidth}
        \multirow{2}{*}{\centering Es} & \multirow{2}{*}{mT5$^{Es}$} & Seen   & 37.89\% & 55.48\% & 62.63\% & 0.654 & 0.779 & 0.825 \\
                                       &        	                  & Unseen & 1.18\%  & 2.44\%  & 3.88\%  & 0.532 & 0.655 & 0.704 \\
        \hline
        \end{tabular}
    }%
    \caption{Comparison of mT5 performance between affective states included and held out from fine-tuning.}
    \label{tab:unseen-emo}
    \vspace{1em}
    \centering
    \small
    \resizebox{0.8\textwidth}{!}{%
        \begin{tabular}{l|rrr|rrr}
            \hline
            Model                       & \multicolumn{1}{l}{Acc@1$\uparrow$} & \multicolumn{1}{l}{Acc@3$\uparrow$} & \multicolumn{1}{l|}{Acc@5$\uparrow$} & \multicolumn{1}{l}{Sim@1$\uparrow$} & \multicolumn{1}{l}{Sim@3$\uparrow$} & \multicolumn{1}{l}{Sim@5$\uparrow$} \\ 
            \hline
            mT5$^{Es}$	 & \textbf{14.07\%} & \textbf{25.31\%} & \textbf{31.37\%} & \textbf{0.462} & \textbf{0.585} & \textbf{0.635} \\
            Llama-3	     & 0.00\%  & 0.00\%  & 0.00\%  & 0.376 & 0.408 & 0.416 \\
            Mixtral	     & 0.04\%  & 0.16\%  & 0.38\%  & 0.342 & 0.358 & 0.372 \\
            \hline
        \end{tabular}
    }%
    \caption{Evaluation of Spanish-fine-tuned mT5, Llama-3, and Mixtral on region-specific Spanish affective states. \textbf{Bolded} metrics highlight the best-performing model.}
    \label{tab:regional-es}
\end{table*}

\subsection{Fixed-Label Set Evaluation}

To evaluate the generalized emotion detection capabilities afforded by fine-tuning on MASIVE, \autoref{tab:fixed-label-english} 
shows the performance of mT5 fine-tuned on existing English and Spanish emotion benchmarks, both with and without prior fine-tuning on MASIVE.
First, when used as a classifier, we find that mT5 fine-tuned on MASIVE first
achieves a higher macro-F1 for all datasets. This suggests that \textbf{fine-tuning on our corpus gives models generalizable knowledge of emotions (Takeaway \#3)}. Because our corpora contain many more affective state labels than the evaluation datasets, models fine-tuned on MASIVE will include more nuanced terms than basic emotions in the top-k predictions. So, as expected, models fine-tuned only on the emotion benchmarks typically achieve higher top-k accuracy and similarity, as they are more likely to predict terms within the smaller label sets. The top-k similarity scores for our models, however, remain high, suggesting that the generated affective states are similar to the ground truth basic emotion labels.

\subsection{Unseen and Regional Set Evaluation}

To analyze how well models generalize beyond affective states explicitly included in fine-tuning, we present performance metrics on seen and unseen affective states in both languages in \autoref{tab:unseen-emo}. In both languages, all models perform considerably better on affective states included in the fine-tuning data than on unseen affective states.
The monolingual T5$^{En}$ model, however, maintains better performance on unseen affective states than mT5$^{En}$, suggesting that monolingual models may 
better generalize to unseen affective states. 

In addition to unseen affective states, we present results on a subset of
Spanish affective states which are region-specific in \autoref{tab:regional-es}. Similarly to results on the full Spanish data, fine-tuned 
mT5$^{Es}$
outperforms both LLMs in top-k accuracy and similarity.
The performance of fine-tuned mT5$^{Es}$ is lower on this regional subset than on the broader set of Spanish texts in MASIVE (\autoref{tab:internal-eval}) but is higher than performance on unseen affective states (\autoref{tab:unseen-emo}).
Llama-3 and Mixtral, which are not fine-tuned on our corpora, also perform significantly worse on the regional subset than they do on the Spanish data as a whole. 
Because top-k accuracy drops significantly on unseen and region-specific affective states (top-k similarity as well, though less so), \textbf{future work in this area should prioritize a generalized understanding of affective states, including regionalisms (Takeaway \#4)}.

\begin{figure*}[ht]
    \centering
    \includegraphics[width=.95\textwidth]{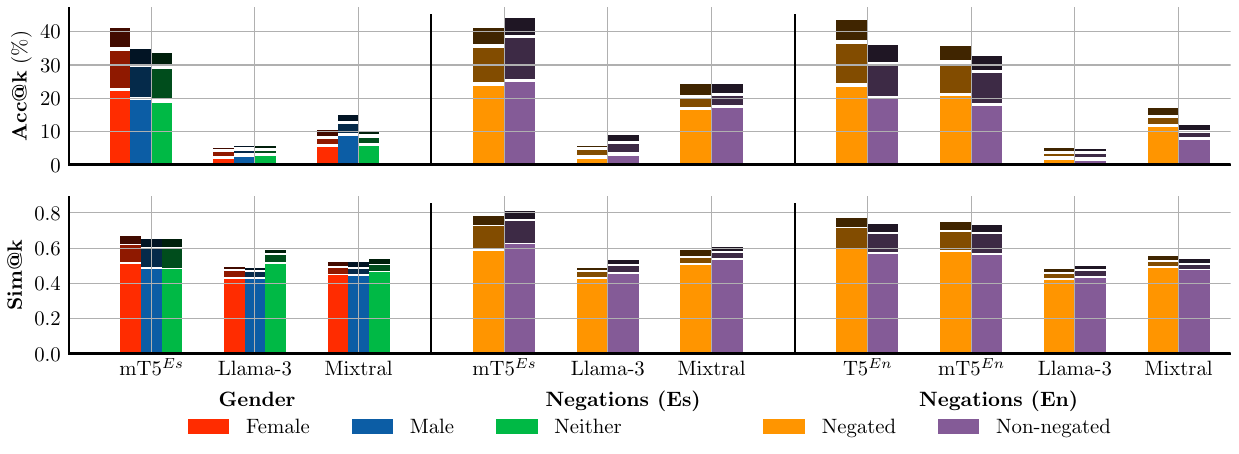}
    \caption{Top-k accuracy and similarity results on subsets reflecting different linguistic constructions in \textsc{MASIVE}: grammatical gender of affective states in Spanish (left) and negated expressions in Spanish (center) and English (right). Shades reflect different values of k separated by small gaps, where the lightest shade represents $k=1$ and the darkest shade represents $k=5$.}
    \label{fig:unique-phen}
\end{figure*}

\begin{table*}[ht]
    \centering
    \small
    \resizebox{\textwidth}{!}{%
        \begin{tabular}{ll|rr|rrr|rrr}
            \hline
            Test Set & Model                    & \multicolumn{1}{l}{NLL$\downarrow$} & \multicolumn{1}{l|}{Log Perp$\downarrow$} & \multicolumn{1}{l}{Acc@1$\uparrow$} & \multicolumn{1}{l}{Acc@3$\uparrow$} & \multicolumn{1}{l|}{Acc@5$\uparrow$} & \multicolumn{1}{l}{Sim@1$\uparrow$} & \multicolumn{1}{l}{Sim@3$\uparrow$} & \multicolumn{1}{l}{Sim@5$\uparrow$} \\ 
            \hline
            \multirow{2}{*}{En}         & mT5$^{En}_S$       & 4.21 & 4.17 & 27.05\% & 41.37\% & 48.71\% & 0.598 & 0.719 & 0.769 \\
                                        & mT5$^{En}_{Tr}$    & 16.86 & 15.79 & 2.18\% & 4.45\% & 6.12\% & 0.418 & 0.532 & 0.579 \\
            \hline
            Es$_{Tr}$	                & mT5$^{Es}$         & 59.37 & 59.23 & 2.35\% & 4.20\% & 5.46\% & 0.369 & 0.448 & 0.482 \\
            \Xhline{4\arrayrulewidth}
            \multirow{2}{*}{Es}	        & mT5$^{Es}$         & 6.91 & 6.89 & 24.51\% & 36.16\% & 41.23\% & 0.610 & 0.734 & 0.781 \\
                                        & mT5$^{Es}_{Tr,S}$	 & 15.80 & 15.54 & 2.37\% & 4.95\% & 6.58\% & 0.378 & 0.472 & 0.517 \\
            \hline
            En$_{Tr}$	                & mT5$^{En}_S$	     & 24.15 & 23.94 & 3.07\% & 6.60\% & 9.39\% & 0.443 & 0.532 & 0.573 \\            
            \hline
        \end{tabular}
    }%
    \caption{Comparison of mT5 fine-tuned on the original data reflecting native language use, fine-tuned on translated data (translate-train), and evaluated on translated data (translate-test) with MASIVE. Only aggregated scores are shown. All fine-tuning sets are randomly subset to the same size as the smallest set, the collected Spanish training set, and results are averaged across 5 different subsets (\textit{n} = 30,958). Models with subsetted data are denoted with $_S$.}
    \label{tab:trans-eval}
\end{table*}

\subsection{Grammatical Gender and Negations}

We break down the top-k accuracy and top-k similarity results for each model by grammatical gender and negations in \autoref{fig:unique-phen}.
We see again that mT5 outperforms both LLMs across all subsets, and that mT5 often places the gold label among the top 3 or 5 predictions if not the top 1. In particular, mT5$^{Es}$ performs better on feminine adjectives than masculine adjectives or those with only a single form, and T5$^{En}$ and mT5$^{En}$ perform better on negated targets than non-negated targets.
Llama-3 and Mixtral achieve highest accuracy for masculine adjectives and highest similarity for single-form adjectives, while for negations, Llama-3 tends to perform better on non-negations and Mixtral tends to perform slightly better on negations. These results suggest that \textbf{explicit training on MASIVE may improve performance specifically on unique features of generative ASI (Takeaway \#5)}.

\subsection{Machine-Translation vs.~Natural Data}

Finally, we evaluate \textit{translate-train} and \textit{translate-test} approaches to cross-lingual transfer \cite{hu-xtreme} against models fine-tuned and evaluated on the original texts.
First, we find an expected drop in performance when models are fine-tuned on machine-translated data for both English and Spanish. 
Interestingly, the average drop in similarity metrics in Spanish (36\%) is notably larger than in English (27\%).
This could perhaps be explained by the translation model performing better 
in the Spanish to English direction than English to Spanish, as well as mT5's ability to better generalize in English than in Spanish.

As an alternative approach to fine-tuning on translated data, we also consider the case where data may be translated at inference time. In these cases (En$_{Tr}$ and Es$_{Tr}$ in \autoref{tab:trans-eval}), we find that performance falls. Artifacts of machine translation have been found to impact evaluation of translation models \citep{freitag-translationese}, and similarly, errors and artifacts of unnatural translation may 
cause
these changes in performance. In contrast to prior work suggesting that performance on the target data translated into English is comparable to fine-tuning on the target language for tasks such as sentiment detection, our results suggest that for our 
task, \textbf{machine-translating the evaluation data leads to poorer performance, and translating either at training or inference time result in similar performance (Takeaway \#6)}.

\section{Related Work}

\paragraph{Emotion Taxonomies. }
Many different models of human emotion have been proposed, intending to capture the universal experience of different emotions across cultures. 
Some of the most notable categorical models in psychology and NLP research are the \citet{ekman-basic,ekman-review} basic emotion set derived from 
facial expression and the \citet{plutchik-emotion} basic emotion set which assumes emotions occur in opposing pairs (e.g. joy and sadness), though other models exist 
(e.g., \citealt{ortony-cognitive, oatley-cognitive, johnson-folk, ps-navarasa}).
Multiple different dimensional models have also been proposed, situating emotions in a space governed by features such as pleasantness and activation \citep{plutchik-emotion,russell-vad,russell-circumplex, bradley-pictures}. Many such models of emotions have been frequently compared and evaluated in psychology and as they apply to emotion detection (see \citealt{rubin-compare,ps-navarasa,lichtenstein-compare}).
    
\paragraph{Emotion and Language Generation. }
Numerous approaches to automated emotion detection in text have been proposed, including emotion lexicons \citep{strapparava-wordnet, staiano-depeche, araque-depeche++, mohammad-nrc} and classification models (see \citealt{acheampoing-review} for a review of approaches). Most of this work focuses on small, finite emotion sets, usually Ekman or Plutchik, though larger sets have been employed, such as those listed in \citet{wang-emotion} and used in other prior work \citep{sintsova-geneva-classification,liew-emotweet-28,subasic-fuzzy-affect,mohammad-fine-grained}. Other works have instead predicted affect dimensions for fine-grained emotion classification \cite{mohammad-semeval}. In contrast, we consider discrete affective states derived from expressions in language and in this work, we evaluate models on predicting over 1,000 distinct affective states. More recently, language generation tasks have been proposed that call for models with greater emotional understanding,
such as emotional dialogue generation \citep{ide-dialogue, song-generation, firdaus-controlled}, controllable generation \citep{goswamy-control, saha-counter}, and emotion trigger summarization \citep{sosea-summarization, zhan-summarization}. 
Given that language generation models have been employed to unify these and other
tasks,
endowing models with a greater understanding of human emotions would greatly benefit multiple applications.

\paragraph{Cross-cultural Emotion Perception. }
Many researchers have suggested that a basic set of emotions are universal, while others have argued that emotions are shaped by culture. Past work has built on Ekman's proposal and provided evidence that emotion categories are universal \citep{ekman-basic, hoemann-universal}, with \citet{sauter-language} finding little support for the argument that language plays a foundational role in perceiving emotions. 
Additionally, past work has
in part supported differences in emotion perception across languages and cultures in humans \citep{chen-culture, mesquita-culture, jackson-semantics,caldwell-bilingual}.
Some work has demonstrated differences
in model performance across languages and cultures
\citep{havaldar-multilingual,hassan-cross}. Others have studied cross-lingual approaches both with and without machine translation on speech tasks \cite{yang-2019-interspeech, rizvi-speaker}, text tasks \cite{patil-overlap, dong-crosslingual}, and specifically for sentiment and emotion tasks \cite{rasooli-et-al-sentiment-transfer,tafreshi-2024-emotion, zhang-crosslingual}. Our work evaluates the use of machine translation, and
we find that machine translation may not be sufficient for cross-lingual transfer on the ASI task.

\section{Conclusion}

In this work, we introduce the novel task of ASI, a language generation task prioritizing the authors' natural expressions of their feelings rather than using a prescribed set of emotion labels. For this task, we automatically collect and publish two datasets of Reddit posts in English and Spanish, both containing over 1,000 unique affective state labels.

We use this dataset to benchmark multilingual generative models, and find that \textbf{(Takeaway \#1)} small fine-tuned T5 and mT5 models outperform zero-shot LLMs. Results specifically show that \textbf{(Takeaway \#2)} T5 outperforms mT5 in English on ASI, suggesting that monolingual models may be more capable. Additionally, we show that \textbf{(Takeaway \#3)} models fine-tuned on our corpora transfer knowledge that generalizes to existing emotion detection benchmarks. In analyzing model performance on unseen affective states and Spanish regionalisms, we argue that \textbf{(Takeaway \#4)} generalization to a broader set of affective states, including those from underrepresented dialects, is an important avenue for future work. With respect to grammatical gender in Spanish and negations, \textbf{(Takeaway \#5)} fine-tuning on MASIVE improves on specific linguistic constructions unique to generative ASI.
Finally, we quantify the observed performance differences when using machine-translated data at fine-tuning or inference time, finding that in contrast to prior work, \textbf{(Takeaway \#6)} machine translation leads to large performance drops.
We hope these results spark future work into ASI to enable prediction of more nuanced feelings in a variety of languages and contexts, and ultimately, enable prediction of an unbounded set of labels. 

\section*{Limitations}
We limit ourselves in this work to investigating two high-resource languages, English and Spanish. We do this in part because, for this application, we find it important that members of the research team be able to speak the languages of study fluently. 
Additionally, we gather data from one source, Reddit, which limits the demographics of the people whose experiences are represented in our data. This choice of data source may particularly limit our Spanish data, which includes fewer texts and labels than English (\autoref{tab:data-stats}). We choose not to control for attributes like topic or subreddit when collecting English and Spanish data separately because we wish to collect a natural variety of data, but this also means that we do not claim our two datasets to be parallel or equivalent.

Our data gathering framework collects only explicit expressions of affective states by searching for statements including an ``I feel''-style template. While we can use models trained on this type of data to predict affective state labels for any input by simply appending an ``I feel'' statement to be filled (see \autoref{sec:external-eval-methods}), our training targets do not include this type of data, and this paradigm impacts the types of affective states we are likely to collect.

We also acknowledge that our choices of specific resources limit our work in various ways. We use only Opus-MT models to perform our machine translation experiments because they exhibit good performance in both languages; however, it is possible that we would see different results with different translation models.
Our similarity metric also uses pre-trained BERT embeddings because of the benefits of contextual embeddings and subword tokenization, but there are many other possible choices of embedding framework that may more accurately capture emotional nuances. Finally, we evaluate only open-source LLMs on our dataset.

\section*{Ethics Statement}

We strictly collect publicly available user-authored texts on the pseudonymous social media website Reddit, but we acknowledge the privacy concerns of users when collecting data from social media. Accordingly, we will release the collected texts only with randomly assigned IDs and usernames stripped. We discourage others from attempting to identify authors of the texts in the collected dataset, and will remove data from the dataset upon request.

Because we rely entirely on open-source models, including open-source LLMs, and make our data available, our results are fully reproducible.  In total, our fine-tuning and evaluation 
amounts to approximately 73 GPU hours using Nvidia A100 GPUs.

Our task allows models to predict a larger set of affective states, capturing more nuanced expressions of an authors' feelings than traditional emotion detection. At the same time, a larger label set could exacerbate the consequences of misclassification in sensitive contexts (e.g., mental health and crisis settings). In some applications of this task where this may be an important consideration, the label set can be artificially restricted, as we show in our external evaluation experiments. 

Finally, the aim of predicting authors' expressions of their own feelings can require models to generate regional or dialectal texts. Prior work has identified dialectal biases in language models (e.g., African American Language; \citealt{deas-aal, groenwold-gen}) and we find that all evaluated models perform poorly on regional varieties of Spanish. We hope future work makes progress toward closing performance gaps among dialects and language varieties.

\bibliography{anthology,custom}

\begin{thebibliography}{67}
\providecommand{\natexlab}[1]{#1}

\bibitem[{Acheampong et~al.(2020)Acheampong, Wenyu, and Nunoo-Mensah}]{acheampoing-review}
Francisca~Adoma Acheampong, Chen Wenyu, and Henry Nunoo-Mensah. 2020.
\newblock \href {https://doi.org/10.1002/eng2.12189} {Text-based emotion detection: Advances, challenges, and opportunities}.
\newblock \emph{Engineering Reports}, 2(7).

\bibitem[{Araque et~al.(2022)Araque, Gatti, Staiano, and Guerini}]{araque-depeche++}
Oscar Araque, Lorenzo Gatti, Jacopo Staiano, and Marco Guerini. 2022.
\newblock \href {https://doi.org/10.1109/TAFFC.2019.2934444} {Depechemood++: A bilingual emotion lexicon built through simple yet powerful techniques}.
\newblock \emph{IEEE Transactions on Affective Computing}, 13(1):496--507.

\bibitem[{Bender and Friedman(2018)}]{bender-data}
Emily~M. Bender and Batya Friedman. 2018.
\newblock \href {https://doi.org/10.1162/tacl_a_00041} {Data statements for natural language processing: Toward mitigating system bias and enabling better science}.
\newblock \emph{Transactions of the Association for Computational Linguistics}, 6:587–604.

\bibitem[{Bird et~al.(2009)Bird, Klein, and Loper}]{bird2009natural}
Steven Bird, Ewan Klein, and Edward Loper. 2009.
\newblock \emph{Natural language processing with Python: analyzing text with the natural language toolkit}.
\newblock " O'Reilly Media, Inc.".

\bibitem[{Bradley et~al.(1992)Bradley, Greenwald, Petry, and Lang}]{bradley-pictures}
Margaret~M. Bradley, Mark~K. Greenwald, Margaret~C. Petry, and Peter~J. Lang. 1992.
\newblock \href {https://doi.org/10.1037/0278-7393.18.2.379} {Remembering pictures: Pleasure and arousal in memory.}
\newblock \emph{Journal of Experimental Psychology: Learning, Memory, and Cognition}, 18(2):379--390.

\bibitem[{Caldwell-Harris and Ay{\c{c}}i{\c{c}}e{\u{g}}i-Dinn(2009)}]{caldwell-bilingual}
Catherine~L. Caldwell-Harris and Ay{\c{s}}e Ay{\c{c}}i{\c{c}}e{\u{g}}i-Dinn. 2009.
\newblock \href {https://doi.org/10.1016/j.ijpsycho.2008.09.006} {Emotion and lying in a non-native language}.
\newblock \emph{International Journal of Psychophysiology}, 71(3):193--204.

\bibitem[{Chen et~al.(2023)Chen, Chung-Fat-Yim, Guo, and Marian}]{chen-culture}
Peiyao Chen, Ashley Chung-Fat-Yim, Taomei Guo, and Viorica Marian. 2023.
\newblock \href {https://doi.org/10.1037/cdp0000577} {Cultural background and input familiarity influence multisensory emotion perception.}
\newblock \emph{Cultural Diversity and Ethnic Minority Psychology}.

\bibitem[{Cowen and Keltner(2017)}]{cowen-distinct}
Alan~S. Cowen and Dacher Keltner. 2017.
\newblock \href {https://doi.org/10.1073/pnas.1702247114} {Self-report captures 27 distinct categories of emotion bridged by continuous gradients}.
\newblock \emph{Proceedings of the National Academy of Sciences}, 114(38).

\bibitem[{Deas et~al.(2023)Deas, Grieser, Kleiner, Patton, Turcan, and McKeown}]{deas-aal}
Nicholas Deas, Jessica Grieser, Shana Kleiner, Desmond Patton, Elsbeth Turcan, and Kathleen McKeown. 2023.
\newblock \href {https://doi.org/10.18653/v1/2023.emnlp-main.421} {Evaluation of {A}frican {A}merican language bias in natural language generation}.
\newblock In \emph{Proceedings of the 2023 Conference on Empirical Methods in Natural Language Processing}, pages 6805--6824, Singapore. Association for Computational Linguistics.

\bibitem[{Demszky et~al.(2020)Demszky, Movshovitz-Attias, Ko, Cowen, Nemade, and Ravi}]{demszky-goemo}
Dorottya Demszky, Dana Movshovitz-Attias, Jeongwoo Ko, Alan Cowen, Gaurav Nemade, and Sujith Ravi. 2020.
\newblock \href {https://doi.org/10.18653/v1/2020.acl-main.372} {{G}o{E}motions: A dataset of fine-grained emotions}.
\newblock In \emph{Proceedings of the 58th Annual Meeting of the Association for Computational Linguistics}, pages 4040--4054, Online. Association for Computational Linguistics.

\bibitem[{Devlin et~al.(2019)Devlin, Chang, Lee, and Toutanova}]{devlin-bert}
Jacob Devlin, Ming-Wei Chang, Kenton Lee, and Kristina Toutanova. 2019.
\newblock \href {https://doi.org/10.18653/v1/N19-1423} {{BERT}: Pre-training of deep bidirectional transformers for language understanding}.
\newblock In \emph{Proceedings of the 2019 Conference of the North {A}merican Chapter of the Association for Computational Linguistics: Human Language Technologies, Volume 1 (Long and Short Papers)}, pages 4171--4186, Minneapolis, Minnesota. Association for Computational Linguistics.

\bibitem[{Dhananjaya et~al.(2024)Dhananjaya, Ranathunga, and Jayasena}]{dhananjaya-lexicon}
Vinura Dhananjaya, Surangika Ranathunga, and Sanath Jayasena. 2024.
\newblock \href {https://doi.org/10.1049/cit2.12333} {Lexicon‐based fine‐tuning of multilingual language models for low‐resource language sentiment analysis}.
\newblock \emph{CAAI Transactions on Intelligence Technology}.

\bibitem[{Dong et~al.(2021)Dong, Zhu, Fu, Xu, and de~Melo}]{dong-crosslingual}
Xin Dong, Yaxin Zhu, Zuohui Fu, Dongkuan Xu, and Gerard de~Melo. 2021.
\newblock \href {https://doi.org/10.18653/v1/2021.acl-long.401} {Data augmentation with adversarial training for cross-lingual {NLI}}.
\newblock In \emph{Proceedings of the 59th Annual Meeting of the Association for Computational Linguistics and the 11th International Joint Conference on Natural Language Processing (Volume 1: Long Papers)}, pages 5158--5167, Online. Association for Computational Linguistics.

\bibitem[{Ekman(1984)}]{ekman-basic}
Paul Ekman. 1984.
\newblock Expression and the nature of emotion.
\newblock \emph{Approaches to emotion}, 3(19):344.

\bibitem[{Ekman(2005)}]{ekman-review}
Paul Ekman. 2005.
\newblock \href {https://doi.org/10.1002/0470013494.ch3} {Basic emotions}.
\newblock In \emph{Handbook of Cognition and Emotion}, pages 45--60. John Wiley \& Sons, Ltd.

\bibitem[{Firdaus et~al.(2020)Firdaus, Chauhan, Ekbal, and Bhattacharyya}]{firdaus-controlled}
Mauajama Firdaus, Hardik Chauhan, Asif Ekbal, and Pushpak Bhattacharyya. 2020.
\newblock Emosen: Generating sentiment and emotion controlled responses in a multimodal dialogue system.
\newblock \emph{IEEE Transactions on Affective Computing}, 13(3):1555--1566.

\bibitem[{Freitag et~al.(2020)Freitag, Grangier, and Caswell}]{freitag-translationese}
Markus Freitag, David Grangier, and Isaac Caswell. 2020.
\newblock \href {https://doi.org/10.18653/v1/2020.emnlp-main.5} {{BLEU} might be guilty but references are not innocent}.
\newblock In \emph{Proceedings of the 2020 Conference on Empirical Methods in Natural Language Processing (EMNLP)}, pages 61--71, Online. Association for Computational Linguistics.

\bibitem[{Goswamy et~al.(2020)Goswamy, Singh, Barkati, and Modi}]{goswamy-control}
Tushar Goswamy, Ishika Singh, Ahsan Barkati, and Ashutosh Modi. 2020.
\newblock Adapting a language model for controlled affective text generation.
\newblock In \emph{Proceedings of the 28th international conference on computational linguistics}, pages 2787--2801.

\bibitem[{Groenwold et~al.(2020)Groenwold, Ou, Parekh, Honnavalli, Levy, Mirza, and Wang}]{groenwold-gen}
Sophie Groenwold, Lily Ou, Aesha Parekh, Samhita Honnavalli, Sharon Levy, Diba Mirza, and William~Yang Wang. 2020.
\newblock \href {https://doi.org/10.18653/v1/2020.emnlp-main.473} {Investigating {A}frican-{A}merican {V}ernacular {E}nglish in transformer-based text generation}.
\newblock In \emph{Proceedings of the 2020 Conference on Empirical Methods in Natural Language Processing (EMNLP)}, pages 5877--5883, Online. Association for Computational Linguistics.

\bibitem[{Hassan et~al.(2022)Hassan, Shaar, and Darwish}]{hassan-cross}
Sabit Hassan, Shaden Shaar, and Kareem Darwish. 2022.
\newblock \href {https://aclanthology.org/2022.lrec-1.751} {Cross-lingual emotion detection}.
\newblock In \emph{Proceedings of the Thirteenth Language Resources and Evaluation Conference}, pages 6948--6958, Marseille, France. European Language Resources Association.

\bibitem[{Havaldar et~al.(2023)Havaldar, Singhal, Rai, Liu, Guntuku, and Ungar}]{havaldar-multilingual}
Shreya Havaldar, Bhumika Singhal, Sunny Rai, Langchen Liu, Sharath~Chandra Guntuku, and Lyle Ungar. 2023.
\newblock \href {https://doi.org/10.18653/v1/2023.wassa-1.19} {Multilingual language models are not multicultural: A case study in emotion}.
\newblock In \emph{Proceedings of the 13th Workshop on Computational Approaches to Subjectivity, Sentiment, {\&} Social Media Analysis}, pages 202--214, Toronto, Canada. Association for Computational Linguistics.

\bibitem[{Hoemann et~al.(2019)Hoemann, Crittenden, Msafiri, Liu, Li, Roberson, Ruark, Gendron, and Barrett}]{hoemann-universal}
Katie Hoemann, Alyssa~N. Crittenden, Shani Msafiri, Qiang Liu, Chaojie Li, Debi Roberson, Gregory~A. Ruark, Maria Gendron, and Lisa~Feldman Barrett. 2019.
\newblock \href {https://doi.org/10.1037/emo0000501} {Context facilitates performance on a classic cross-cultural emotion perception task.}
\newblock \emph{Emotion}, 19(7):1292--1313.

\bibitem[{Honnibal et~al.(2020)Honnibal, Montani, Van~Landeghem, and Boyd}]{honnibal2020spacy}
Matthew Honnibal, Ines Montani, Sofie Van~Landeghem, and Adriane Boyd. 2020.
\newblock \href {https://doi.org/10.5281/zenodo.1212303} {{spaCy: Industrial-strength Natural Language Processing in Python}}.

\bibitem[{Hu et~al.(2020)Hu, Ruder, Siddhant, Neubig, Firat, and Johnson}]{hu-xtreme}
Junjie Hu, Sebastian Ruder, Aditya Siddhant, Graham Neubig, Orhan Firat, and Melvin Johnson. 2020.
\newblock Xtreme: A massively multilingual multi-task benchmark for evaluating cross-lingual generalisation.
\newblock In \emph{International Conference on Machine Learning}, pages 4411--4421. PMLR.

\bibitem[{Ide and Kawahara(2021)}]{ide-dialogue}
Tatsuya Ide and Daisuke Kawahara. 2021.
\newblock \href {https://doi.org/10.18653/v1/2021.naacl-srw.15} {Multi-task learning of generation and classification for emotion-aware dialogue response generation}.
\newblock In \emph{Proceedings of the 2021 Conference of the North American Chapter of the Association for Computational Linguistics: Student Research Workshop}, pages 119--125, Online. Association for Computational Linguistics.

\bibitem[{Isbister et~al.(2021)Isbister, Carlsson, and Sahlgren}]{isbister-translation}
Tim Isbister, Fredrik Carlsson, and Magnus Sahlgren. 2021.
\newblock \href {https://aclanthology.org/2021.nodalida-main.42} {Should we stop training more monolingual models, and simply use machine translation instead?}
\newblock In \emph{Proceedings of the 23rd Nordic Conference on Computational Linguistics (NoDaLiDa)}, pages 385--390, Reykjavik, Iceland (Online). Link{\"o}ping University Electronic Press, Sweden.

\bibitem[{Jackson et~al.(2019)Jackson, Watts, Henry, List, Forkel, Mucha, Greenhill, Gray, and Lindquist}]{jackson-semantics}
Joshua~Conrad Jackson, Joseph Watts, Teague~R. Henry, Johann-Mattis List, Robert Forkel, Peter~J. Mucha, Simon~J. Greenhill, Russell~D. Gray, and Kristen~A. Lindquist. 2019.
\newblock \href {https://doi.org/10.1126/science.aaw8160} {Emotion semantics show both cultural variation and universal structure}.
\newblock \emph{Science}, 366(6472):1517--1522.

\bibitem[{Jiang et~al.(2024)Jiang, Sablayrolles, Roux, Mensch, Savary, Bamford, Chaplot, de~las Casas, Hanna, Bressand, Lengyel, Bour, Lample, Lavaud, Saulnier, Lachaux, Stock, Subramanian, Yang, Antoniak, Scao, Gervet, Lavril, Wang, Lacroix, and Sayed}]{jiang-mixtral}
Albert~Q. Jiang, Alexandre Sablayrolles, Antoine Roux, Arthur Mensch, Blanche Savary, Chris Bamford, Devendra~Singh Chaplot, Diego de~las Casas, Emma~Bou Hanna, Florian Bressand, Gianna Lengyel, Guillaume Bour, Guillaume Lample, Lélio~Renard Lavaud, Lucile Saulnier, Marie-Anne Lachaux, Pierre Stock, Sandeep Subramanian, Sophia Yang, Szymon Antoniak, Teven~Le Scao, Théophile Gervet, Thibaut Lavril, Thomas Wang, Timothée Lacroix, and William~El Sayed. 2024.
\newblock \href {https://arxiv.org/abs/2401.04088} {Mixtral of experts}.
\newblock \emph{arXiv preprint}.

\bibitem[{Johnson-Laird and Oatley(1998)}]{johnson-folk}
Philip~N Johnson-Laird and Keith Oatley. 1998.
\newblock Basic emotions, rationality, and folk theory.
\newblock In \emph{Consciousness and Emotion in Cognitive Science}, pages 289--311. Routledge.

\bibitem[{Kathunia et~al.(2024)Kathunia, Kaif, Arora, and Narotam}]{kathunia-translation}
Aekansh Kathunia, Mohammad Kaif, Nalin Arora, and N~Narotam. 2024.
\newblock \href {https://arxiv.org/abs/2405.02887} {Sentiment analysis across languages: Evaluation before and after machine translation to english}.
\newblock \emph{Preprint}, arXiv:2405.02887.

\bibitem[{Lichtenstein et~al.(2008)Lichtenstein, Oehme, Kupschick, and J\"{u}rgensohn}]{lichtenstein-compare}
Antje Lichtenstein, Astrid Oehme, Stefan Kupschick, and Thomas J\"{u}rgensohn. 2008.
\newblock \href {https://doi.org/10.1007/978-3-540-85099-1_4} {Comparing two emotion models for deriving affective states from physiological data}.
\newblock In \emph{Affect and Emotion in Human-Computer Interaction}, pages 35--50. Springer Berlin Heidelberg.

\bibitem[{Liew et~al.(2016)Liew, Turtle, and Liddy}]{liew-emotweet-28}
Jasy Suet~Yan Liew, Howard~R. Turtle, and Elizabeth~D. Liddy. 2016.
\newblock \href {https://aclanthology.org/L16-1183} {{E}mo{T}weet-28: A fine-grained emotion corpus for sentiment analysis}.
\newblock In \emph{Proceedings of the Tenth International Conference on Language Resources and Evaluation ({LREC}'16)}, pages 1149--1156, Portoro{\v{z}}, Slovenia. European Language Resources Association (ELRA).

\bibitem[{Mesquita et~al.(2016)Mesquita, Boiger, and Leersnyder}]{mesquita-culture}
Batja Mesquita, Michael Boiger, and Jozefien~De Leersnyder. 2016.
\newblock \href {https://doi.org/10.1016/j.copsyc.2015.09.015} {The cultural construction of emotions}.
\newblock \emph{Current Opinion in Psychology}, 8:31--36.

\bibitem[{Mohammad et~al.(2018)Mohammad, Bravo-Marquez, Salameh, and Kiritchenko}]{mohammad-semeval}
Saif Mohammad, Felipe Bravo-Marquez, Mohammad Salameh, and Svetlana Kiritchenko. 2018.
\newblock \href {https://doi.org/10.18653/v1/S18-1001} {{S}em{E}val-2018 task 1: Affect in tweets}.
\newblock In \emph{Proceedings of the 12th International Workshop on Semantic Evaluation}, pages 1--17, New Orleans, Louisiana. Association for Computational Linguistics.

\bibitem[{Mohammad and Turney(2010)}]{mohammad-nrc}
Saif Mohammad and Peter Turney. 2010.
\newblock \href {https://aclanthology.org/W10-0204} {Emotions evoked by common words and phrases: Using {M}echanical {T}urk to create an emotion lexicon}.
\newblock In \emph{Proceedings of the {NAACL} {HLT} 2010 Workshop on Computational Approaches to Analysis and Generation of Emotion in Text}, pages 26--34, Los Angeles, CA. Association for Computational Linguistics.

\bibitem[{Mohammad and Kiritchenko(2015)}]{mohammad-fine-grained}
Saif~M. Mohammad and Svetlana Kiritchenko. 2015.
\newblock \href {https://doi.org/10.1111/coin.12024} {Using hashtags to capture fine emotion categories from tweets}.
\newblock \emph{Computational Intelligence}, 31(2):301--326.

\bibitem[{Oatley and Johnson-laird(1987)}]{oatley-cognitive}
Keith Oatley and P.~N. Johnson-laird. 1987.
\newblock \href {https://doi.org/10.1080/02699938708408362} {Towards a cognitive theory of emotions}.
\newblock \emph{Cognition \&: Emotion}, 1(1):29--50.

\bibitem[{Ortony et~al.(1988)Ortony, Clore, and Collins}]{ortony-cognitive}
Andrew Ortony, Gerald~L. Clore, and Allan Collins. 1988.
\newblock \href {https://doi.org/10.1017/cbo9780511571299} {\emph{The Cognitive Structure of Emotions}}.
\newblock Cambridge University Press.

\bibitem[{Patil et~al.(2022)Patil, Talukdar, and Sarawagi}]{patil-overlap}
Vaidehi Patil, Partha Talukdar, and Sunita Sarawagi. 2022.
\newblock \href {https://doi.org/10.18653/v1/2022.acl-long.18} {Overlap-based vocabulary generation improves cross-lingual transfer among related languages}.
\newblock In \emph{Proceedings of the 60th Annual Meeting of the Association for Computational Linguistics (Volume 1: Long Papers)}, pages 219--233, Dublin, Ireland. Association for Computational Linguistics.

\bibitem[{Plaza-del Arco et~al.(2024)Plaza-del Arco, Cercas~Curry, Cercas~Curry, and Hovy}]{plaza-del-arco-etal-2024-emotion-analysis}
Flor~Miriam Plaza-del Arco, Alba~A. Cercas~Curry, Amanda Cercas~Curry, and Dirk Hovy. 2024.
\newblock \href {https://aclanthology.org/2024.lrec-main.506} {Emotion analysis in {NLP}: Trends, gaps and roadmap for future directions}.
\newblock In \emph{Proceedings of the 2024 Joint International Conference on Computational Linguistics, Language Resources and Evaluation (LREC-COLING 2024)}, pages 5696--5710, Torino, Italia. ELRA and ICCL.

\bibitem[{Plaza~del Arco et~al.(2020)Plaza~del Arco, Strapparava, Urena~Lopez, and Martin}]{plaza-del-arco-emoevent}
Flor~Miriam Plaza~del Arco, Carlo Strapparava, L.~Alfonso Urena~Lopez, and Maite Martin. 2020.
\newblock \href {https://aclanthology.org/2020.lrec-1.186} {{E}mo{E}vent: A multilingual emotion corpus based on different events}.
\newblock In \emph{Proceedings of the Twelfth Language Resources and Evaluation Conference}, pages 1492--1498, Marseille, France. European Language Resources Association.

\bibitem[{Plutchik(1980)}]{plutchik-emotion}
Robert Plutchik. 1980.
\newblock A general psychoevolutionary theory of emotion.
\newblock In \emph{Theories of emotion}, pages 3--33. Elsevier.

\bibitem[{PS and Mahalakshmi(2017)}]{ps-navarasa}
Sreeja PS and G~Mahalakshmi. 2017.
\newblock Emotion models: a review.
\newblock \emph{International Journal of Control Theory and Applications}, 10(8):651--657.

\bibitem[{Raffel et~al.(2019)Raffel, Shazeer, Roberts, Lee, Narang, Matena, Zhou, Li, and Liu}]{raffel-t5}
Colin Raffel, Noam Shazeer, Adam Roberts, Katherine Lee, Sharan Narang, Michael Matena, Yanqi Zhou, Wei Li, and Peter~J. Liu. 2019.
\newblock \href {https://arxiv.org/abs/1910.10683} {Exploring the limits of transfer learning with a unified text-to-text transformer}.
\newblock \emph{CoRR}, abs/1910.10683.

\bibitem[{Rasooli et~al.(2018)Rasooli, Farra, Radeva, Yu, and McKeown}]{rasooli-et-al-sentiment-transfer}
Mohammad~Sadegh Rasooli, Noura Farra, Axinia Radeva, Tao Yu, and Kathleen McKeown. 2018.
\newblock \href {http://www.jstor.org/stable/44988391} {Cross-lingual sentiment transfer with limited resources}.
\newblock \emph{Machine Translation}, 32(1/2):143--165.

\bibitem[{Rizvi et~al.(2023)Rizvi, Jamatia, Rudrapal, Chakma, and Gamb{\"a}ck}]{rizvi-speaker}
Amaan Rizvi, Anupam Jamatia, Dwijen Rudrapal, Kunal Chakma, and Bj{\"o}rn Gamb{\"a}ck. 2023.
\newblock \href {https://aclanthology.org/2023.ranlp-1.105} {Cross-lingual speaker identification for {I}ndian languages}.
\newblock In \emph{Proceedings of the 14th International Conference on Recent Advances in Natural Language Processing}, pages 979--987, Varna, Bulgaria. INCOMA Ltd., Shoumen, Bulgaria.

\bibitem[{Rubin and Talarico(2009)}]{rubin-compare}
David~C. Rubin and Jennifer~M. Talarico. 2009.
\newblock \href {https://doi.org/10.1080/09658210903130764} {A comparison of dimensional models of emotion: Evidence from emotions, prototypical events, autobiographical memories, and words}.
\newblock \emph{Memory}, 17(8):802--808.

\bibitem[{Russell(1980)}]{russell-circumplex}
James~A. Russell. 1980.
\newblock \href {https://doi.org/10.1037/h0077714} {A circumplex model of affect.}
\newblock \emph{Journal of Personality and Social Psychology}, 39(6):1161--1178.

\bibitem[{Russell and Mehrabian(1977)}]{russell-vad}
James~A Russell and Albert Mehrabian. 1977.
\newblock Evidence for a three-factor theory of emotions.
\newblock \emph{Journal of research in Personality}, 11(3):273--294.

\bibitem[{Saha et~al.(2022)Saha, Singh, Kumar, Mathew, and Mukherjee}]{saha-counter}
Punyajoy Saha, Kanishk Singh, Adarsh Kumar, Binny Mathew, and Animesh Mukherjee. 2022.
\newblock \href {https://doi.org/10.24963/ijcai.2022/716} {Countergedi: A controllable approach to generate polite, detoxified and emotional counterspeech}.
\newblock In \emph{Proceedings of the Thirty-First International Joint Conference on Artificial Intelligence}, IJCAI-2022, page 5157–5163. International Joint Conferences on Artificial Intelligence Organization.

\bibitem[{Sauter(2018)}]{sauter-language}
Disa~A. Sauter. 2018.
\newblock \href {https://doi.org/10.1177/1754073917693924} {Is there a role for language in emotion perception?}
\newblock \emph{Emotion Review}, 10(2):111--115.

\bibitem[{Shazeer and Stern(2018)}]{shazeer2018adafactor}
Noam Shazeer and Mitchell Stern. 2018.
\newblock \href {https://proceedings.mlr.press/v80/shazeer18a.html} {Adafactor: Adaptive learning rates with sublinear memory cost}.
\newblock In \emph{Proceedings of the 35th International Conference on Machine Learning}, volume~80 of \emph{Proceedings of Machine Learning Research}, pages 4596--4604. PMLR.

\bibitem[{Sintsova et~al.(2013)Sintsova, Musat, and Pu}]{sintsova-geneva-classification}
Valentina Sintsova, Claudiu Musat, and Pearl Pu. 2013.
\newblock \href {https://aclanthology.org/W13-1603} {Fine-grained emotion recognition in olympic tweets based on human computation}.
\newblock In \emph{Proceedings of the 4th Workshop on Computational Approaches to Subjectivity, Sentiment and Social Media Analysis}, pages 12--20, Atlanta, Georgia. Association for Computational Linguistics.

\bibitem[{Song et~al.(2019)Song, Zheng, Liu, Xu, and Huang}]{song-generation}
Zhenqiao Song, Xiaoqing Zheng, Lu~Liu, Mu~Xu, and Xuanjing Huang. 2019.
\newblock \href {https://doi.org/10.18653/v1/P19-1359} {Generating responses with a specific emotion in dialog}.
\newblock In \emph{Proceedings of the 57th Annual Meeting of the Association for Computational Linguistics}, pages 3685--3695, Florence, Italy. Association for Computational Linguistics.

\bibitem[{Sosea et~al.(2023)Sosea, Zhan, Li, and Caragea}]{sosea-summarization}
Tiberiu Sosea, Hongli Zhan, Junyi~Jessy Li, and Cornelia Caragea. 2023.
\newblock \href {https://doi.org/10.18653/v1/2023.acl-long.531} {Unsupervised extractive summarization of emotion triggers}.
\newblock In \emph{Proceedings of the 61st Annual Meeting of the Association for Computational Linguistics (Volume 1: Long Papers)}, pages 9550--9569, Toronto, Canada. Association for Computational Linguistics.

\bibitem[{Staiano and Guerini(2014)}]{staiano-depeche}
Jacopo Staiano and Marco Guerini. 2014.
\newblock \href {https://doi.org/10.3115/v1/P14-2070} {Depeche mood: a lexicon for emotion analysis from crowd annotated news}.
\newblock In \emph{Proceedings of the 52nd Annual Meeting of the Association for Computational Linguistics (Volume 2: Short Papers)}, pages 427--433, Baltimore, Maryland. Association for Computational Linguistics.

\bibitem[{Strapparava and Valitutti(2004)}]{strapparava-wordnet}
Carlo Strapparava and Alessandro Valitutti. 2004.
\newblock \href {http://www.lrec-conf.org/proceedings/lrec2004/pdf/369.pdf} {{W}ord{N}et affect: an affective extension of {W}ord{N}et}.
\newblock In \emph{Proceedings of the Fourth International Conference on Language Resources and Evaluation ({LREC}{'}04)}, Lisbon, Portugal. European Language Resources Association (ELRA).

\bibitem[{Subasic and Huettner(2001)}]{subasic-fuzzy-affect}
P.~Subasic and A.~Huettner. 2001.
\newblock \href {https://doi.org/10.1109/91.940962} {Affect analysis of text using fuzzy semantic typing}.
\newblock \emph{IEEE Transactions on Fuzzy Systems}, 9(4):483--496.

\bibitem[{Tafreshi et~al.(2024)Tafreshi, Vatsal, and Diab}]{tafreshi-2024-emotion}
Shabnam Tafreshi, Shubham Vatsal, and Mona Diab. 2024.
\newblock \href {https://arxiv.org/abs/2402.18424} {Emotion classification in low and moderate resource languages}.
\newblock \emph{Preprint}, arXiv:2402.18424.

\bibitem[{Tiedemann and Thottingal(2020)}]{tiedmann-opusmt}
J{\"o}rg Tiedemann and Santhosh Thottingal. 2020.
\newblock {OPUS-MT} — {B}uilding open translation services for the {W}orld.
\newblock In \emph{Proceedings of the 22nd Annual Conferenec of the European Association for Machine Translation (EAMT)}, Lisbon, Portugal.

\bibitem[{VandenBos(2007)}]{vandenbos-apa}
Gary~R VandenBos. 2007.
\newblock \emph{APA dictionary of psychology.}
\newblock American Psychological Association.

\bibitem[{Wang et~al.(2020)Wang, Ho, and Cambria}]{wang-emotion}
Zhaoxia Wang, Seng-Beng Ho, and Erik Cambria. 2020.
\newblock \href {https://doi.org/10.1007/s11042-019-08328-z} {A review of emotion sensing: categorization models and algorithms}.
\newblock \emph{Multimedia Tools and Applications}, 79(47–48):35553–35582.

\bibitem[{Xue et~al.(2021{\natexlab{a}})Xue, Constant, Roberts, Kale, Al-Rfou, Siddhant, Barua, and Raffel}]{xue-etal-2021-mt5}
Linting Xue, Noah Constant, Adam Roberts, Mihir Kale, Rami Al-Rfou, Aditya Siddhant, Aditya Barua, and Colin Raffel. 2021{\natexlab{a}}.
\newblock \href {https://doi.org/10.18653/v1/2021.naacl-main.41} {m{T}5: A massively multilingual pre-trained text-to-text transformer}.
\newblock In \emph{Proceedings of the 2021 Conference of the North American Chapter of the Association for Computational Linguistics: Human Language Technologies}, pages 483--498, Online. Association for Computational Linguistics.

\bibitem[{Xue et~al.(2021{\natexlab{b}})Xue, Constant, Roberts, Kale, Al-Rfou, Siddhant, Barua, and Raffel}]{xue-mt5}
Linting Xue, Noah Constant, Adam Roberts, Mihir Kale, Rami Al-Rfou, Aditya Siddhant, Aditya Barua, and Colin Raffel. 2021{\natexlab{b}}.
\newblock \href {https://doi.org/10.18653/v1/2021.naacl-main.41} {m{T}5: A massively multilingual pre-trained text-to-text transformer}.
\newblock In \emph{Proceedings of the 2021 Conference of the North American Chapter of the Association for Computational Linguistics: Human Language Technologies}, pages 483--498, Online. Association for Computational Linguistics.

\bibitem[{Yang and Hirschberg(2019)}]{yang-2019-interspeech}
Zixiaofan Yang and Julia Hirschberg. 2019.
\newblock \href {https://api.semanticscholar.org/CorpusID:202891380} {Linguistically-informed training of acoustic word embeddings for low-resource languages}.
\newblock In \emph{Interspeech}.

\bibitem[{Zhan et~al.(2022)Zhan, Sosea, Caragea, and Li}]{zhan-summarization}
Hongli Zhan, Tiberiu Sosea, Cornelia Caragea, and Junyi~Jessy Li. 2022.
\newblock \href {https://doi.org/10.18653/v1/2022.emnlp-main.642} {Why do you feel this way? summarizing triggers of emotions in social media posts}.
\newblock In \emph{Proceedings of the 2022 Conference on Empirical Methods in Natural Language Processing}, pages 9436--9453, Abu Dhabi, United Arab Emirates. Association for Computational Linguistics.

\bibitem[{Zhang et~al.(2024)Zhang, Zhao, Zhang, Zhao, and Bao}]{zhang-crosslingual}
Jinghui Zhang, Yuan Zhao, Siqin Zhang, Ruijing Zhao, and Siyu Bao. 2024.
\newblock \href {https://aclanthology.org/2024.wassa-1.53} {Enhancing cross-lingual emotion detection with data augmentation and token-label mapping}.
\newblock In \emph{Proceedings of the 14th Workshop on Computational Approaches to Subjectivity, Sentiment, {\&} Social Media Analysis}, pages 528--533, Bangkok, Thailand. Association for Computational Linguistics.

\end{thebibliography}

\appendix

\section{Data Statement} 
\label{app:data-statement}

\subsection{Curation Rationale}

The aim of collecting the texts contained in MASIVE was to produce both a training dataset and benchmark for affective state identification. Affective state identification tasks models with predicting individual terms reflecting how a text's author feels, and in particular, predicting terms that would be used by the author. The dataset collection process was designed to automatically extract a large set of possible affective state labels from texts where an author explicitly describes how they feel. Both an English and Spanish version of the dataset were collected in the same fashion to enable research on cross-lingual work, as well as a small set of regional Spanish to enable work on linguistic variation. We intend to make the dataset publicly available

\subsection{Language Variety}

MASIVE contains texts both in English (en) and Spanish (es). Data collection was not restricted to a particular variety of English or Spanish, and distributions of these varieties likely reflects the overall demographics of English and Spanish-speaking users on Reddit. A small set of data was collected specifically to reflect Spanish specific to particular regions, including terms primarily associated with Spanish spoken in Mexico, Spain, Venezuela, and El Salvador among other regions and countries.

\subsection{Annotator Demographics}

Two sets of annotators were involved in validating the automatically extracted labels in MASIVE. For the English data, annotators were 2 native English-speakers and Psychology undergraduate students. 
Both English data annotators were American and female.
For the Spanish data, annotators were 2 native Spanish-speakers and graduate students in the department of Latin American and Iberian Cultures. 
The Spanish data annotators were Colombian and Ecuadorian, and both were male.

\subsection{Speech Situation}

The collected texts in MASIVE were not restricted to a particular range of time, and may have been published anytime between the founding of Reddit (2005) and the time of data collection (April, 2024). Texts were also not restricted to a particular place, but likely reflect the countries of origin of English and Spanish-speaking Reddit users. All texts were originally written and published on Reddit, which may or may not have been edited before they were included in the dataset. As with most interactions through Reddit posts, the texts reflect asynchronous interactions and are likely intended for a general public audience in most cases.

\subsection{Text Characteristics}

The texts in MASIVE may discuss a wide variety of topics. All texts, however, contain explicit expressions of feelings or explicit mentions of terms that may reflect feelings. Thus, many texts may reflect personal narratives that provide context for an author's feelings. Thus, the dataset may also discuss sensitive topics and include the kinds of offensive or harmful content that can be found online.

\section{Data Examples} \label{app:data-ex}

Examples of input texts and the identified affective state labels from MASIVE are included in \autoref{tab:data-exs}.

\begin{table*}[!htbp]
    \centering
    \small
    \begin{tabular}{P{7cm}|P{7cm}}
        \hline
        En & Es \\
        \hline
        Four wait-lists to start my cycle... is this normal or a bad sign? Applied to all my schools in mid-late December.  Over the past two weeks I've received my first decisions: wait-lists from Penn, Michigan, Cornell, and UT.  I'm feeling {\color{red} discouraged} at the moment because I definitely wasn't expecting to be waitlisted from all four of these schools. 7sage predictor gave me a 49\%, 74\%, 83\%, and 93\% chance at these schools, respectively.  Getting waitlisted at a target, likely target, and two safeties to start my cycle has me fairly concerned.  I had two strong LOR's (fantastic relationship with two of my college professors), and thought I did a great job on my P.S. I'm wondering if my lack of internship/work experience is going to hurt me this cycle, as I've been a server/front desk staff for two years since graduating college. Would appreciate anyone's thoughts on what might be going on/what to expect. I blanketed the T14 as well as applying to UCLA, BU, and Vandy. Thanks in advance for any guidance!! Wishing good luck to all of you! & Le platique a "un amigo" de una chava y la agrego a sus redes sociales Me causa inseguridad y molestia que haya hecho eso ¿con que intención lo hizo? El me había dicho que no la conocía ni ubicaba y ahora de la nada ya la tiene en redes. Ella y yo no tenemos nada serio, apenas estamos empezando a salir. Lo siento como una traición a mi confianza y falta de respeto, estoy pensando en confrontarlo y dejar de hablarle ya que eso no lo hace los amigos, y yo tengo que darme mi lugar y no permitir esas faltas de respeto. ¿Estoy exagerando? Necesito opiniones, no sé si este exagerando, mi desconfianza me este haciendo una jugada o lo estoy viendo como debería ser. No sé, no estoy seguro, pero internamente si me siento {\color{red} enojado} y {\color{red} molesto}.\\
        \hline
        Why I just started the SHEIN \$100 game and I was .05 cents away and they said I reached the max assets. It said if I got people to join I’d receive 1 point for each and I only needed 5 and it started giving me .05 cents then said game over. I was at the \$100 if it followed what it said but the game did not. What a rip!!! Not happy I feel {\color{red} cheated} for sure! You make millions SHEIN, why do that to your customers?! & [Serio] me vaciaron la caja de ahorro Eso, hace una hora y media tenía 4k y ahora -4 pesos, lo único que hice fue sacar 500 pesos y después hace una compra con débito y en farmacity, no sé que hacer, estoy {\color{red} desesperada}, no me atienden del número de banco nación, que puedo hacer? Ayuda por favor!! UPDATE: después de sufrir dos horas volvió todo mi saldo así como se fue, vacíe la cuenta y cambie de nuevo el pin de la tarjeta, ahora tengo un miedo de que me pase de vuelta, que puedo hacer para prevenir??
    \end{tabular}
    \caption{Example texts and accompanying affective state labels from the English and Spanish subsets of MASIVE. Affective state labels are colored {\color{red} red}.}
    \label{tab:data-exs}
\end{table*}

\section{Data Collection and Annotation} \label{app:data}

\subsection{Seed Emotions} \label{app:seed-emos}

The specific adjective forms of the Ekman emotions used to seed our bootstrapping procedure are shown in \autoref{tab:seed-emos}. These are also the terms used as the gold in our fixed-set label evaluation, with the addition of `\textit{nothing}' for the no-emotion class if it is used.

For fixed-label evaluation of GoEmotions (27), the following terms are used for the expanded label set: `\textit{admiration}', `\textit{amused}', `\textit{angry}', `\textit{annoyed}', `\textit{approving}', `\textit{caring}', `\textit{confused}', `\textit{curious}', `\textit{desire}', `\textit{disappointed}', `\textit{disapproval}', `\textit{disgusted}', `\textit{embarrassed}', `\textit{excited}', `\textit{afraid}', `\textit{grateful}', `\textit{grief}', `\textit{happy}', '\textit{love}', `\textit{nervous}', `\textit{optimistic}', `\textit{proud}', `\textit{realized}', `\textit{relieved}', `\textit{remorseful}', `\textit{sad}', `\textit{surprised}', and `\textit{nothing}'.

\begin{table}[ht]
    \centering
    \small
    \begin{tabular}{ P{1.5cm} P{1.5cm} | P{1.5cm} P{1.5cm}}
         \hline
         \multicolumn{2}{c|}{\centering En} & \multicolumn{2}{c}{\centering Es} \\ 
         \hline
         happy      & surprised     & feliz     & sorprendido \\
         sad        & disgusted     & triste    & desagradado \\
         angry      & afraid        & enojado   & asustado \\
         \hline
    \end{tabular}
    \caption{Seed emotions (Ekman) for each language used in collecting MASIVE.}
    \label{tab:seed-emos}
\end{table}

\subsection{Regional Spanish Affective States}
\label{app:regional}

To collect affective state labels associated with one or more particular Spanish-speaking regions, we use the following set of terms:
`\textit{mamado/a}', `\textit{patitieso/a}', `\textit{emputado/a}', `\textit{encandilado/a}', `\textit{arrechado/a}', `\textit{fastidiado/a}', `\textit{encabronado/a}', `\textit{hallado/a}', `\textit{rayado/a}', `\textit{achispado/a}', `\textit{ahuevado/a}', `\textit{enrabiado/a}', `\textit{tusa}', `\textit{chocho/a}', `\textit{encachimbado/a}', `\textit{bravo/a}', `\textit{apantallado/a}', `\textit{embromado/a}', `\textit{engorilado/a}', `\textit{alicaido/a}', `\textit{flipando/a}', `\textit{cagado/a}', `\textit{aguitado/a}', `\textit{engrinchado/a}', `\textit{chato/a}', `\textit{chipil}', `\textit{picado/a}', `\textit{bajoneado/a}', `\textit{acojonado/a}', `\textit{arrecho/a}'"

The terms are not exhaustive, but reflect varieties of Spanish spoken in Spain, Chile, Colombia, Venezuela, Mexico, Bolivia, Argentina, Uruguay, and Paraguay.

\subsection{Data Annotation} \label{app:data-annot}

\begin{figure*}[ht]
    \includegraphics[width=\textwidth]{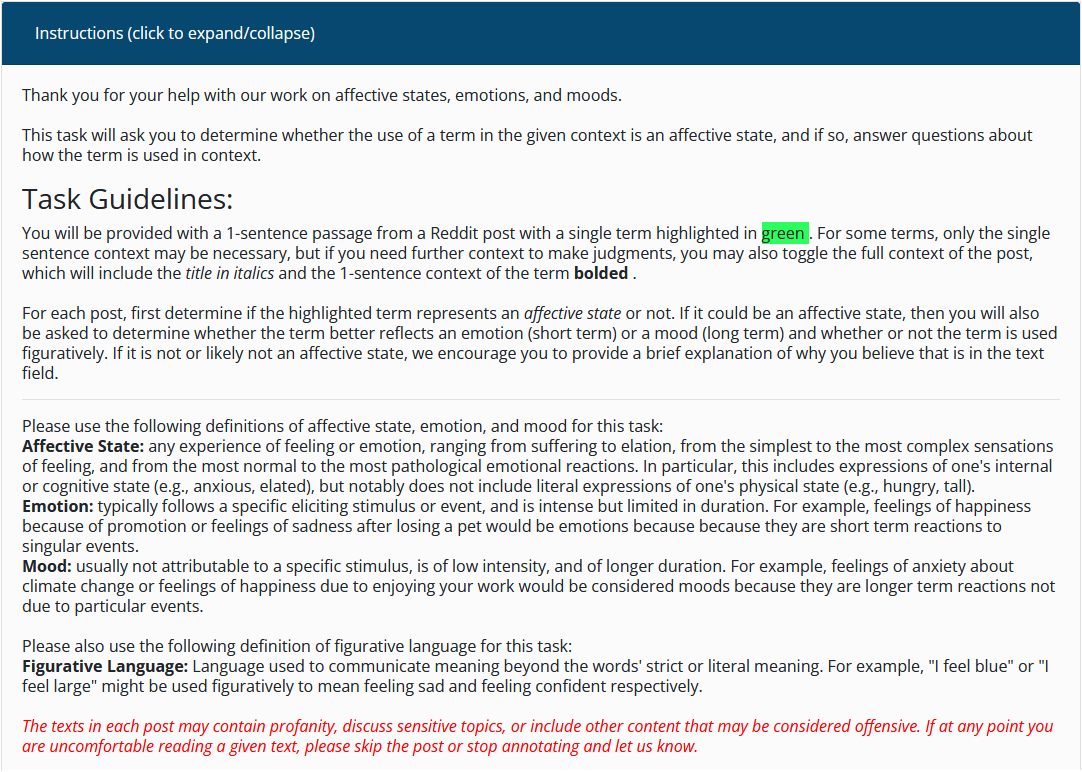}
    \caption{Instructions provided to our human annotators, including definitions. Annotators may collapse or expand the instructions at will.}
    \label{fig:ann-instructions}
\end{figure*}

\begin{figure*}[ht]
    \includegraphics[width=\textwidth]{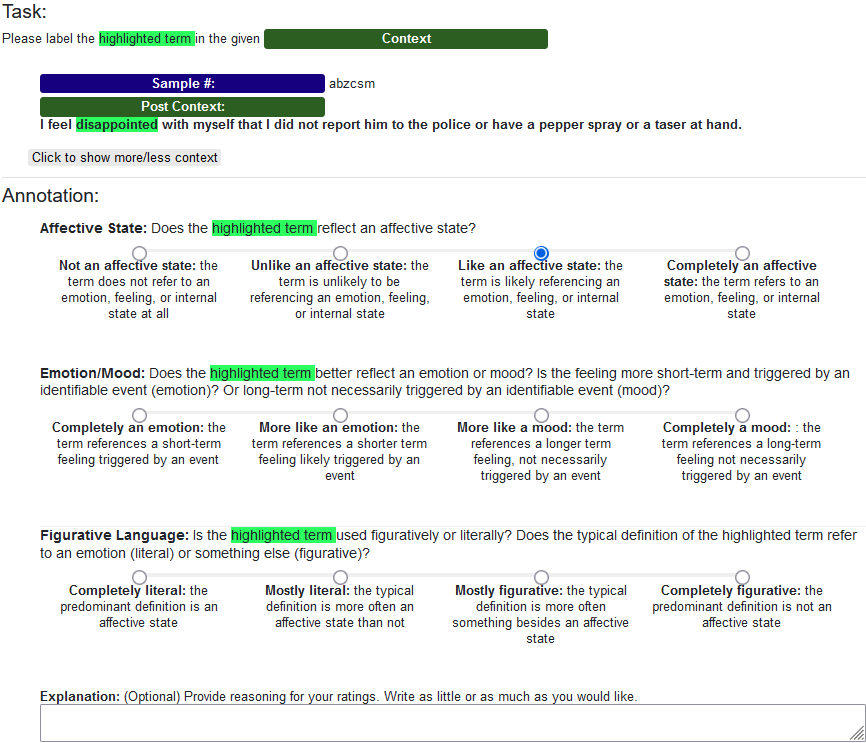}
    \caption{Human annotation interface with a sample datapoint. Clicking the button to show more or less context toggles the display of the full Reddit post vs. the one-sentence context. As shown, the Emotion/Mood and Figurative Language questions only appear if the highlighted term is judged \textbf{like an affective state} or \textbf{completely an affective state}.}
    \label{fig:ann-interface}
\end{figure*}

The instructions and interface given to our human annotators are shown in \autoref{fig:ann-instructions} and \autoref{fig:ann-interface}, respectively. 
Annotators were paid \$23/hour for their work in accordance with the standards of their university. Each annotator completed a pilot task of 30 examples before beginning to annotate the data in order to build familiarity with the platform and task. 

\section{Experimental Setup} \label{app:methods}

\subsection{Generation Configuration} \label{app:gen-config}

\paragraph{Checkpoints. }
Throughout our experiments, we use the large variants of T5 (770 million parameters; \textit{google-t5/t5-large}) and mT5 (1.2  billion parameters; \textit{google/mt5-large}). For our two LLMs, we evaluate the instruct variants of Llama-3 (8 billion parameters; \textit{meta-llama/Meta-Llama-3-8B-Instruct}) loaded in bfloat16 and Mixtral (7$\times$22 billion parameters; \textit{mistralai/Mixtral-8x22B-Instruct-v0.1}). Mixtral is accessed through the \url{fireworks.ai} API.

Beyond the evaluated models, we use two open-source, unidirectional translation models for our translation experiments. In particular, we employ the Helsinki-NLP English-to-Spanish (\textit{Helsinki-NLP/opus-mt-en-es}) and Spanish-to-English (\textit{Helsinki-NLP/opus-mt-es-en}) models. We also use a multilingual BERT checkpoint as part of the similarity metric (168 million parameters; \textit{bert-base-multilingual-uncased}). Finally, we also rely on \texttt{spacy} \citep{honnibal2020spacy} to identify parts of speech in English (\textit{en\_core\_web\_md}) and Spanish (\textit{es\_core\_news\_md}) during our data collection.

\paragraph{Generation.}
For T5, mT5, and Llama-3, we use beam search to generate the top-k most likely predictions, with 5 beams (as we need only the top-5 outputs). We use the default settings of Huggingface's \href{https://huggingface.co/docs/transformers/en/main_classes/text_generation#transformers.GenerationConfig}{\texttt{GenerationConfig}}, including, e.g., no repetition penalty, etc.; though we expect a single-word output, we allow generations of up to 32 tokens.

The API used to run inference with Mixtral does not allow retrieving the top 5 most probable predictions as we do with the aforementioned models. Instead, Mixtral predictions are generated with a top-k of 5, and a temperature of 0.5. The top 5 candidate generations are then reranked by the log-probability according to Mixtral to be used in evaluating the ranked, top-5 predictions. Also due to accessing Mixtral through an API, we were not able to calculate the log perplexity of the ground truth labels.

\paragraph{Hyperparameters.} T5 and mT5 models are fine-tuned with a batch size of 4 for 3 epochs each. Model parameters are optimized using Adafactor \citep{shazeer2018adafactor} as implemented by Huggingface's \texttt{transformers} with a learning rate of $1 \times 10^{-4}$, Huggingface's linear learning rate scheduler with default parameters, and a weight decay parameter (here, an L2 penalty) of 0.01. For each model, all data is tokenized using the correct pre-trained tokenizer corresponding to its pre-trained checkpoint. Any input that is longer than 512 tokens (including the end-of-sequence token) is trimmed to fit; in order to preserve the target affective state masks and the grammatical integrity of the text, this trimming removes full sentences (as parsed by \texttt{nltk}; \citealt{bird2009natural}) from the end of the text if possible (i.e., if this will not remove a target mask), or the beginning otherwise, until the text fits within 512 tokens.

\subsection{Prompts} \label{app:prompts}

\autoref{tab:prompts} shows the prompts provided to Mixtral and Llama-3 throughout our experiments. In a minority of cases, models would reply in the form "Here is a list of terms to fill each <MASK>: ", in which case, only the terms following the colon were considered as the model's prediction.

\begin{table}[ht]
    \centering
    \small
    \begin{tabular}{P{1cm}|P{5.5cm}}
        \hline
        Lang & Prompt \\
        \hline
        En   & Determine the most likely term reflecting a feeling to replace each <MASK> in the following text: "\textit{<MASKED\_POST>}" Provide a single emotion term for each <MASK> token. Do not introduce the answer, respond ONLY with a comma-separated list of lowercase terms:   \\
        \hline
        Es   & Determine the most likely term reflecting a feeling to replace each <MASK> in the following text: "\textit{<MASKED\_POST>}" Provide a single emotion term for each <MASK> token. Do not introduce the answer, respond ONLY with a comma-separated list of lowercase terms in Spanish: \\
        \hline
    \end{tabular}
    \caption{Prompts provided to Llama-3 and Mixtral for evaluation. At inference time, \textit{<POST>} is replaced with the input text containing masked affective states.}
    \label{tab:prompts}
\end{table}

\subsection{Machine Translation Configuration}
In the fine-tuning experiment, we subset the English data and translated English-to-Spanish data to keep the number of training steps constant across settings.  
For these two models, we repeat the experiment with 5 different random subsets and report the averages across the five trials.

\section{Top-K Similarity} \label{app:topk-sim}
Let $P = [p_1, p_2, p_3, ...p_n]$, where $n \geq k$,
be a list of predictions ordered according to descending likelihood, and let $g$ be the gold (where $p_i$ and $g$ are strings). Additionally, let $E(x)$ be a function on a term $x$ that incorporates 100 tokens of context,
tokenizes and embeds the sequence with a pre-trained BERT tokenizer, and returns the contextual embedding corresponding to the first sub-word token in $x$. Then, we report top-k similarity specifically as
\begin{equation*}
    \text{sim}_k(P, g)=\max_{i \leq k} \big[\text{cosine\_sim}(E(p_i), E(g))\big]
\end{equation*} 

\section{Extended Results} \label{app:results}

\subsection{Limited Evaluation for Llama-3} \label{app:llama-subset}

For some inputs, Llama-3 would decline to make a prediction, particularly for inputs that discuss topics such as depression or drug use. While these are important topics for models to be able to accurately analyze as they are increasingly applied in mental health contexts, Llama-3's behavior may unfairly skew its evaluation results. \autoref{tab:llama-reeval} presents updated results for Llama-3 on the subset of texts for which the model's response followed the correct format. 60\% of English, 70\% of Spanish, and 76\% of regional Spanish responses by Llama-3 were formatted correctly. Across datasets, scores improve only by up to $\sim$2.4\% top-k accuracy and $\sim$.04 top-k similarity. Considering these results, no conclusions made are altered.

\begin{table*}
    \centering
    \small
    \resizebox{\textwidth}{!}{%
        \begin{tabular}{l|r|rrr|rrr}
            \hline
            Lang                       & \multicolumn{1}{l}{\% Valid} & \multicolumn{1}{l}{Acc@1$\uparrow$} & \multicolumn{1}{l}{Acc@3$\uparrow$} & \multicolumn{1}{l|}{Acc@5$\uparrow$} & \multicolumn{1}{l}{Sim@1$\uparrow$} & \multicolumn{1}{l}{Sim@3$\uparrow$} & \multicolumn{1}{l}{Sim@5$\uparrow$} \\ 
            \hline
            En          & 59.69\% & 2.05\% & 3.63\% & 4.69\% & 0.433 & 0.479 & 0.502 \\
            Es          & 69.51\% & 3.53\% & 6.57\% & 8.27\% & 0.467 & 0.516 & 0.541 \\
            Es (Reg)    & 76.03\% & 0.00\% & 0.00\% & 0.00\% & 0.384 & 0.425 & 0.436 \\
            \hline
        \end{tabular}
    }%
    \caption{Evaluation results of Llama-3 on each MASIVE dataset only considering samples with correctly formatted responses of the form "\textit{prediction\_1}, \textit{prediction\_2}, etc..."}
    \label{tab:llama-reeval}
\end{table*}

\subsection{Full Fixed-Label Set Results}
Extended results from the fixed-label evaluation are given in \autoref{tab:full-fixed-label}. Notably, we include results using T5 in English, where T5 represents a model fine-tuned only on the target dataset and $\text{T5}^{MAS}$ represents a model fine-tuned on MASIVE and then fine-tuned on the target dataset. Precision, recall, and F1 are calculated by ranking the adjective forms of each emotion class (\autoref{app:data}) according to model likelihood and taking the most likely one as the predicted class, while top-k accuracy and similarity are calculated in a generative setting as in the remainder of the paper. T5 generally scores well on F1; pre-training on MASIVE does not usually improve T5's performance on GoEmotions, while it does for EmoEvent (En).

\begin{table*}[ht]
    \adjustbox{max width=\textwidth}{%
    \centering
    \small
    \begin{tabular}{l|l|rrr|lll|lll}
        \hline
        Dataset                                          & Model & \multicolumn{1}{c}{P} & \multicolumn{1}{c}{R} & \multicolumn{1}{c}{F1} & Acc@1& Acc@3 & Acc@5 & Sim@1 & Sim@3 & Sim@5 \\ 
        \hline
        \multicolumn{1}{l|}{\multirow{4}{*}{GoEmotions (7)}}  
                                                    & T5$^{En}$	     & \textbf{56.77} & 24.67 & 21.09 & \textbf{38.53\%} & 47.95\% & 55.74\% & 0.734 & 0.775 & 0.804 \\
                                                    & T5$^{MAS}$	 & 31.10 & 24.07 & 19.67 & 34.50\% & 47.26\% & 55.09\% & 0.708 & 0.774 & 0.803 \\
                                                    & mT5$^{En}$	 & 33.63 & 19.28 & 16.25 & 38.49\% & \textbf{70.73\%} & \textbf{85.99\%} & \textbf{0.736} & \textbf{0.884} & \textbf{0.946} \\
                                                    & mT5$^{MAS}$	 & 33.06 & \textbf{39.81} & \textbf{28.30} & 17.49\% & 32.11\% & 39.25\% & 0.629 & 0.733 & 0.771 \\
        \hline
        \multicolumn{1}{l|}{\multirow{4}{*}{GoEmotions (27)}} 
                                                    & T5$^{En}$	     & 23.31 & 13.05 & 9.73  & 2.03\% & 3.27\%  & 3.86\%  & 0.197 & 0.461 & 0.492 \\
                                                    & T5$^{MAS}$	 & 11.09 & 5.19  & 1.26  & 2.64\% & 4.10\%  & 5.14\%  & 0.506 & 0.560 & 0.574 \\
                                                    & mT5$^{En}$	 & 12.57 & 4.77  & 2.24  & 2.53\% & \textbf{12.90\%} & \textbf{23.51\%} & \textbf{0.525} & \textbf{0.614} & \textbf{0.670} \\
                                                    & mT5$^{MAS}$	 & \textbf{27.08} & \textbf{18.76} & \textbf{11.92} & \textbf{7.54\%} & 12.22\% & 15.16\% & 0.508 & 0.602 & 0.639 \\
        \hline
        \multicolumn{1}{l|}{\multirow{4}{*}{EmoEvent (En)}}   
                                                    & T5$^{En}$	     & 51.06 & 26.63 & 25.05 & 27.44\% & 51.30\% & 56.38\% & 0.541 & 0.777 & 0.811 \\
                                                    & T5$^{MAS}$	 & \textbf{35.46} & \textbf{33.65} & 28.08 & 32.55\% & 59.56\% & 71.85\% & 0.671 & 0.837 & 0.892 \\
                                                    & mT5$^{En}$	 & 30.06 & 14.36 & 2.84  & 10.50\% & \textbf{71.70\%} & \textbf{93.64\%} & 0.630 & \textbf{0.880} & \textbf{0.974} \\
                                                    & mT5$^{MAS}$	 & 34.81 & 32.74 & \textbf{29.55} & \textbf{33.40\%} & 57.06\% & 69.38\% & \textbf{0.712} & 0.842 & 0.893 \\
        \Xhline{4\arrayrulewidth}
        \multicolumn{1}{l|}{\multirow{2}{*}{EmoEvent (Es)}}	  
                                                    & mT5$^{Es}$	 & 26.13 & 14.52 & 6.41 & 24.29\% & 70.34\% & \textbf{89.12\%} & 0.713 & 0.882 & \textbf{0.955} \\
                                                    & mT5$^{MAS}$	 & \textbf{54.93} & \textbf{21.54} & \textbf{17.80} & \textbf{39.75\%} & \textbf{82.62\%} & 86.11\% & \textbf{0.750} & \textbf{0.918} & 0.935 \\
        \hline
    \end{tabular}
    }%
    \caption{Fixed-label evaluation of our models on prior emotion classification datasets. The best performance under each metric for each dataset is \textbf{bolded}.}
    \label{tab:full-fixed-label}
\end{table*}

\end{document}